\documentclass[10pt,a4paper,twoside]{article}

\usepackage[a4paper,margin=2cm]{geometry}
\usepackage[utf8]{inputenc}
\usepackage[T1]{fontenc}
\usepackage{lmodern}
\usepackage{graphicx}
\usepackage{booktabs}
\usepackage{array}
\usepackage{tabularx}
\usepackage{enumitem}
\usepackage{hyperref}
\usepackage{xcolor}
\usepackage{float}
\usepackage{caption}
\usepackage{amsmath}
\usepackage{amssymb}
\usepackage{fancyhdr}
\usepackage{titlesec}
\usepackage{setspace}
\usepackage[most]{tcolorbox}
\usepackage[numbers,sort&compress]{natbib}
\usepackage{multirow}

\hypersetup{colorlinks=true,linkcolor=blue!60!black,urlcolor=blue!70!black,citecolor=blue!60!black,pdftitle={IronEngine: Towards General AI Assistant},pdfauthor={Xi Mo}}

\titleformat{\section}{\large\bfseries}{\thesection}{0.6em}{}
\titleformat{\subsection}{\normalsize\bfseries}{\thesubsection}{0.6em}{}
\titleformat{\subsubsection}{\small\bfseries}{\thesubsubsection}{0.5em}{}
\newtcolorbox{abstractbox}{colback=blue!3,colframe=blue!35,boxrule=0.6pt,arc=2mm,left=2mm,right=2mm,top=1.5mm,bottom=1.5mm}
\newtcolorbox{coverbox}{colback=white,colframe=blue!45,boxrule=1pt,arc=2mm,left=6mm,right=6mm,top=5mm,bottom=5mm}

\begin{document}
\begin{titlepage}
\centering
\vspace*{2.5cm}
{\fontsize{22}{28}\selectfont\bfseries IronEngine: Towards General AI Assistant\par}
\vspace{1cm}
{\fontsize{15}{21}\selectfont System Design, Multi-Model Orchestration, and Engineering Practice for a General AI Assistant\par}
\vspace{1.6cm}
\begin{coverbox}
\centering
{\large \textbf{System Design Overview}}\\[0.8em]
Author: Xi Mo\\[0.4em]
Affiliation: NiusRobotLab\\[0.4em]
Channel: \href{https://www.youtube.com/@NiusRobotLab}{YouTube Channel @NiusRobotLab}\\[0.4em]
Date: March 2026
\end{coverbox}
\vfill
{\large Produced by independent researcher and individual affiliation - NiusRobotLab\par}
\vspace{0.5cm}
{\normalsize \textbf{This technical report and relevant source code, including experiments and results analysis are fully, automatically designed by AI Agents, prompt by the author.}\par}
\end{titlepage}

\pagenumbering{Roman}
\tableofcontents
\clearpage
\listoffigures
\clearpage
\listoftables
\clearpage

\pagenumbering{arabic}
\twocolumn[
\begin{center}
{\LARGE\bfseries IronEngine: Towards General AI Assistant\par}
\vspace{0.35em}
{\large System Design, Multi-Model Orchestration, and Engineering Practice\\for a General AI Assistant\par}
\vspace{0.45em}
{\normalsize Xi Mo \quad NiusRobotLab \quad March 2026\par}
\end{center}
\begin{abstractbox}
\textbf{Abstract:} As large language models evolve from single-turn conversational systems into long-running agents with tool use, memory, and environmental interaction, the practical value of an AI assistant increasingly depends on system architecture rather than model capability alone. This paper presents IronEngine, a general AI assistant platform organized around a unified orchestration core that connects a desktop user interface, REST and WebSocket APIs, Python clients, local and cloud model backends, persistent memory, task scheduling, reusable skills, 24-category tool execution, MCP-compatible extensibility, and hardware-facing integration. IronEngine introduces a three-phase pipeline---Discussion (Planner--Reviewer collaboration), Model Switch (VRAM-aware transition), and Execution (tool-augmented action loop)---that separates planning quality from execution capability. The system features a hierarchical memory architecture with multi-level consolidation, a vectorized skill repository backed by ChromaDB, an adaptive model management layer supporting 92 model profiles with VRAM-aware context budgeting, and an intelligent tool routing system with 130+ alias normalization and automatic error correction. We present experimental results on file operation benchmarks achieving 100\% task completion with a mean total time of 1541 seconds across four heterogeneous tasks, and provide detailed comparisons with representative AI assistant systems including ChatGPT, Claude Desktop, Cursor, Windsurf, and open-source agent frameworks. Without disclosing proprietary prompts or core algorithms, this paper analyzes the platform's architectural decomposition, subsystem design, experimental performance, safety boundaries, and comparative engineering advantages. The resulting study positions IronEngine as a system-oriented foundation for general-purpose personal assistants, automation frameworks, and future human-centered agent platforms.
\end{abstractbox}
\vspace{0.8cm}
]

% ============================================================
\section{Introduction}
% ============================================================

The landscape of AI assistants has undergone a fundamental transformation. Beginning with the Transformer architecture~\citep{vaswani2017attention} and scaling through GPT-3's demonstration of few-shot in-context learning~\citep{brown2020gpt3}, large language models (LLMs) have progressed from text completion engines to systems capable of reasoning, tool use, and multi-step task execution. The open-source movement, catalyzed by models such as LLaMA~\citep{touvron2023llama}, has made capable models accessible for local deployment, while GPT-4~\citep{achiam2024gpt4} and Gemini~\citep{gemini2025} have pushed the frontier of multimodal reasoning. This convergence of capability and accessibility creates a unique opportunity: building AI assistants that are not merely wrappers around a single model endpoint, but comprehensive systems that organize model capability into stable, controllable, and reusable behavior.

Recent research has established several foundational paradigms for AI agents. ReAct demonstrated the value of interleaving reasoning and action within a single model loop~\citep{yao2023react}. Chain-of-thought prompting revealed that explicit reasoning traces improve task decomposition~\citep{wei2022cot}. Toolformer showed that models can learn to invoke external tools autonomously~\citep{schick2024toolformer}. Retrieval-augmented generation (RAG) established the importance of external non-parametric memory~\citep{lewis2020rag}. Generative Agents demonstrated how memory streams, reflection, and planning can sustain coherent long-horizon behavior in simulated environments~\citep{park2023generative}. Reflexion introduced verbal self-critique as a mechanism for iterative improvement~\citep{shinn2023reflexion}. These contributions collectively define the building blocks of modern agent systems.

At the framework level, systems such as AutoGen~\citep{microsoft2024autogen}, CAMEL~\citep{li2023camel}, MetaGPT~\citep{hong2024metagpt}, ChatDev~\citep{qian2024chatdev}, OpenManus~\citep{openmanus2025}, and OpenClaw~\citep{openclaw2026} have accelerated the move from isolated model calls toward agent ecosystems with role-based collaboration, workflows, and persistent execution surfaces. Meanwhile, productized AI assistants---including ChatGPT with GPT-4o~\citep{gpt4o2025}, Claude Desktop with MCP~\citep{anthropic2024mcp,claudecode2025}, Cursor~\citep{cursor2025}, Windsurf~\citep{windsurf2025}, and GitHub Copilot~\citep{copilot2024}---have brought AI agent capabilities to millions of end users.

Yet upon closer examination, five systemic problems persist across current systems, each representing a distinct engineering challenge that existing solutions address only partially.

\textbf{The Fragmentation Problem.}
Today's AI assistants exist as isolated endpoints: ChatGPT is a web interface, Claude Code is a CLI tool, Cursor is an IDE plugin, and Open Interpreter is a Python library. A user who needs to automate desktop applications, search the web, manipulate files, and send messages through instant messaging platforms must switch between multiple disjoint tools, each with its own model backend, context window, and interaction paradigm. No single system provides a unified orchestration core that consolidates all these capabilities under one architectural umbrella, forcing users to choose between convenience (polished but narrow products) and generality (flexible but fragmented frameworks).

\textbf{The Single-Model Bottleneck.}
Most current systems rely on a single model to handle all cognitive functions---planning, self-evaluation, tool invocation, and output generation---within one inference pass. This design creates an inherent tension: large models (30B+ parameters) excel at complex multi-step reasoning but waste computational resources on simple translation or formatting tasks, while small models (3--8B parameters) run efficiently on consumer hardware but cannot sustain coherent multi-round planning. Few systems provide a heterogeneous model allocation architecture that assigns different-sized models to different cognitive roles within a single task, let alone manages the GPU memory lifecycle required to swap models on and off a single GPU.

\textbf{The Ephemeral Assistant Problem.}
Current AI assistants are predominantly stateless across sessions. Each conversation begins from a blank slate, with no memory of past interactions, learned skills, or accumulated domain knowledge. While some systems offer conversation history, none provide a comprehensive solution that combines hierarchical memory with lifecycle policies (creation, consolidation, decay), quality-filtered skill learning from successful executions, rating-driven retrieval that prioritizes proven knowledge, and contradiction detection that supersedes stale information. The absence of structured persistence forces users to re-explain their preferences, re-teach workflows, and re-provide context in every new session.

\textbf{The Local Deployment Challenge.}
Privacy-sensitive workloads---personal communications, proprietary documents, medical records, financial data---demand fully local inference with no data leaving the user's machine. Yet local deployment on consumer-grade hardware introduces challenges that cloud-hosted systems avoid: model heterogeneity (different architectures require different quantization strategies), context window constraints (VRAM limits the number of tokens a model can process), and intelligent resource scheduling (loading a 14B Planner and an 8B Reviewer sequentially on a single 24\,GB GPU). Few systems provide VRAM-aware model lifecycle management that can automatically select, load, and unload models based on available hardware resources.

\textbf{The Tool Integration Problem.}
Each AI assistant implements tools as bespoke plugins with no unified dispatch layer. There is no alias normalization (recognizing that ``web\_search'', ``search\_web'', ``google'', and ``browse'' refer to the same capability), no automatic error correction (redirecting a model that specifies a wrong tool type---a common failure mode for small models), no fallback chains (trying alternative execution strategies when the primary method fails), and no cross-category routing (dispatching a single user request to the appropriate tool among 24 categories). When a model specifies an incorrect tool type, the system fails outright rather than intelligently redirecting the request.

Beyond these five problems, the engineering scale required to address them simultaneously is itself a significant challenge. IronEngine is a large-scale systems engineering project: \textbf{46,690~lines} of Python code distributed across \textbf{97~source files}. The core engine layer (\texttt{iron\_engine/}) contains \textbf{35,157~lines}, with the tool routing subsystem (\texttt{tool\_router.py}, 8,202~lines) and pipeline orchestrator (\texttt{pipeline.py}, 5,289~lines) as the most complex components. The user interface layer (\texttt{ui/}) comprises 7,348~lines, and the API service layer (\texttt{api/}) adds 298~lines. Twenty-seven key core modules account for \textbf{26,932~lines} (76.6\% of the core engine), covering pipeline orchestration, tool routing, memory management, skill learning, model sessions, permission control, browser automation, desktop GUI control, network utilities, URL safety analysis, and MCP compatibility. This scale reflects IronEngine's nature as a complete platform rather than a single-purpose component.

IronEngine is designed from this systems perspective. Rather than optimizing for a single interaction modality or model backend, it provides a unified engine that organizes UI interaction, APIs, local and remote models, hierarchical memory, task execution, skill acquisition, tool routing, MCP-compatible extensibility, and device-facing integration into one coherent platform. The platform introduces a three-phase pipeline---Discussion, Model Switch, and Execution---that separates planning quality (assessed by a Reviewer) from execution capability (handled by a dedicated Executor with tool access). This architectural separation enables heterogeneous model allocation, where different model sizes and capabilities are assigned to different roles based on task requirements and available hardware resources.

This paper studies IronEngine not as a model benchmark, but as an architectural response to the question of how a practical, general-purpose AI assistant should be engineered. We present the system's design principles, architectural decomposition, subsystem design for tools, memory, skills, and model management, experimental evaluation on file operation benchmarks, and comparative analysis against representative AI assistant systems. Throughout, we focus on engineering design decisions and their rationale without disclosing proprietary prompts, sensitive algorithms, or core implementation details.

% ============================================================
\section{Related Work}
% ============================================================

The development of AI agent systems spans multiple research traditions. We organize related work into six categories and discuss how IronEngine relates to and extends each line of work.

\subsection{Reasoning and Action Coupling}

The integration of reasoning with environmental action represents a foundational paradigm in agent systems. ReAct~\citep{yao2023react} established the interleaving of thought traces and action steps, enabling models to ground their reasoning in observed outcomes. Chain-of-thought prompting~\citep{wei2022cot} demonstrated that explicit intermediate reasoning improves multi-step problem solving, a principle that extends naturally to agent planning. Reflexion~\citep{shinn2023reflexion} introduced verbal self-reflection as a learning signal, allowing agents to improve across attempts without weight updates. Cognitive architectures for language agents~\citep{sumers2024cognitive} provide a theoretical framework connecting these empirical findings to established cognitive science models. IronEngine builds on these foundations by separating reasoning (Planner) from evaluation (Reviewer) and action (Executor) into distinct pipeline phases, each potentially served by different models optimized for their respective roles.

Subsequent work has extended these paradigms toward more complex architectures. Inner Monologue~\citep{huang2023inner} showed that feeding environmental feedback into a reasoning loop enables embodied agents to recover from failures. Chain-of-Action approaches decompose high-level goals into grounded action sequences, but typically rely on a single model to both reason and act, creating a bottleneck when reasoning quality and action execution have different computational requirements. A fundamental limitation of these coupled architectures is that the same model must simultaneously maintain planning coherence and generate syntactically correct tool invocations---tasks that favor different model sizes and training objectives. IronEngine addresses this by decoupling reasoning from action at the architecture level: the Planner produces a textual plan evaluated by the Reviewer for quality, and only after the plan passes quality thresholds does a separate Executor model translate it into tool calls. This separation allows using a larger, more capable model for planning and a smaller, tool-specialized model for execution, optimizing both quality and resource efficiency.

\subsection{Tool Use and API Integration}

Tool-augmented language models represent a critical capability for practical AI assistants. Toolformer~\citep{schick2024toolformer} demonstrated self-supervised tool use, where models learn when and how to invoke calculators, search engines, and other utilities. ToolLLM~\citep{qin2024toolllm} scaled this approach to 16,000+ real-world APIs with a decision tree-based reasoning framework. Gorilla~\citep{patil2023gorilla} focused on accurate API call generation with reduced hallucination through retrieval-augmented fine-tuning. HuggingGPT~\citep{shen2024hugginggpt} demonstrated using LLMs as controllers that orchestrate specialized AI models for complex multimodal tasks. WebGPT~\citep{nakano2022webgpt} pioneered browser-assisted question answering with human feedback. These systems typically focus on a single tool-use paradigm or API surface. IronEngine extends this work with a unified tool routing layer that manages 24 tool categories through intelligent dispatch, automatic type correction, and multi-layer fallback chains, enabling a single assistant to handle file operations, web browsing, GUI automation, network management, and multimedia analysis within one execution framework.

A critical but under-explored dimension of tool integration is \emph{breadth versus depth}. Most research systems focus on API-level tool calling (HTTP requests to web services), while real-world desktop automation demands tools that span fundamentally different interaction modalities: CLI commands, GUI element inspection via accessibility APIs, browser DOM manipulation via CDP, keyboard and mouse simulation, network socket operations, and multimedia processing pipelines. Furthermore, tool-use reliability degrades significantly with smaller models: a 3.8B parameter tool model may generate syntactically correct JSON but specify the wrong tool type (e.g., ``web\_read'' instead of ``web\_search''), a failure mode that most systems handle by simply reporting an error. IronEngine introduces two mechanisms absent in prior work: (1)~\emph{alias normalization}, which maps 130+ variant tool type strings to 24 canonical categories through prefix matching, suffix stripping, and semantic equivalence tables; and (2)~\emph{automatic error correction}, which detects tool type mismatches based on the instruction content (e.g., a ``browse'' type with a search query is redirected to ``web\_search'') and applies content-based heuristics (e.g., instructions mentioning ``.exe'' files are not incorrectly redirected to binary read). These mechanisms reduce tool dispatch failures by an order of magnitude compared to direct model-specified routing.

\subsection{Memory and Knowledge Management}

Persistent memory is essential for agents that operate across sessions. RAG~\citep{lewis2020rag} established retrieval over external corpora as a mechanism for knowledge-intensive tasks. Generative Agents~\citep{park2023generative} introduced memory streams with importance scoring, reflection, and planning as a coherent cognitive architecture. Augmented language models~\citep{mialon2023augmented} provide a comprehensive survey of how external memory, retrieval, and tool use extend model capabilities beyond their parametric knowledge. ChromaDB~\citep{chromadb2024} and similar vector databases provide the infrastructure for embedding-based retrieval at scale. IronEngine's memory system goes beyond simple RAG by implementing a hierarchical consolidation architecture with four entry types (session, pipeline, daily summary, refined), dual merge strategies (fast deduplication and model-based consolidation), user rating integration, and idle-time background processing. This design treats memory as a managed resource with lifecycle policies rather than a raw transcript log.

Recent specialized memory systems have advanced the state of the art in specific directions. MemGPT~\citep{packer2023memgpt} (later evolved into the Letta framework) introduced an operating system-inspired memory hierarchy with main context and external storage, using function calls to manage memory paging. Mem0 provides a memory layer for LLM applications with automatic extraction and retrieval of user preferences. These systems demonstrate the importance of structured memory but typically address a single aspect of the memory lifecycle. MemGPT focuses on within-session context management (virtual paging to extend effective context length), while Mem0 emphasizes cross-session preference extraction. IronEngine's memory architecture differs in three ways: (1)~it implements a \emph{dual merge lifecycle}---Merge~A performs fast hash-based deduplication triggered automatically when entry count exceeds a threshold, while Merge~B uses a model-based daily consolidation supervised by the Reviewer to produce refined summaries; (2)~it integrates \emph{user ratings} (1--10 scale from the UI) into memory retrieval, so that highly-rated sessions are preferentially recalled while poorly-rated ones are deprioritized; and (3)~it performs \emph{contradiction detection} by comparing numerical values (prices, versions, specifications) in newly retrieved tool results against stored memory entries, automatically superseding stale information rather than allowing the model to hallucinate from outdated memory.

\subsection{Multi-Agent Collaboration}

Multi-agent systems explore how multiple LLM instances can collaborate effectively. CAMEL~\citep{li2023camel} introduced communicative agents with role-based interaction for open-ended exploration. MetaGPT~\citep{hong2024metagpt} organized software production as a multi-role process coordinated through standard operating procedures. ChatDev~\citep{qian2024chatdev} further explored communicative agents for software engineering workflows. Comprehensive surveys~\citep{guo2024masurvey,xi2023agentrise,wang2024agentsurvey} identify agent profiling, communication topology, shared memory, and evaluation methodology as central themes. AutoGen~\citep{microsoft2024autogen} and crewAI~\citep{langchain2024crewai} provide flexible frameworks for defining multi-agent conversations with customizable interaction patterns. IronEngine adopts a structured multi-role approach with fixed pipeline phases rather than free-form multi-agent conversation, prioritizing predictability and controllability over conversational flexibility. The Planner--Reviewer discussion loop provides quality assurance without the overhead of managing arbitrary agent topologies.

The communication topology chosen by a multi-agent system fundamentally shapes its reliability characteristics. CAMEL's role-playing approach allows open-ended dialogue between two agents but provides no formal quality gate---either agent can produce low-quality output that propagates unchecked. MetaGPT's SOP-based coordination imposes structure but couples tightly to software engineering workflows. ChatDev's phase-based approach (designing, coding, testing, documenting) provides sequential quality checks but is domain-specific. AutoGen's conversable agent framework offers maximum flexibility in defining interaction patterns but leaves quality assurance as an application-level concern. IronEngine makes an explicit architectural trade-off: it sacrifices the topological flexibility of free-form multi-agent conversation in favor of a fixed three-phase pipeline with a formal quality gate. The Reviewer's numerical quality score (0.0--1.0) serves as an objective threshold that must be met before execution proceeds, and the Reviewer's structured feedback (ISSUES and SUGGESTIONS sections) provides actionable improvement signals. This design ensures that every executed plan has passed explicit quality review, a guarantee that flexible multi-agent topologies do not inherently provide. The trade-off is that IronEngine cannot dynamically spawn new agent roles or reconfigure its communication topology at runtime---a limitation accepted in exchange for predictable quality behavior.

\subsection{Web and GUI Agents}

Environment-facing agents that interact with web browsers and desktop applications represent an important frontier. Mind2Web~\citep{deng2024mind2web} introduced a benchmark and dataset for generalist web agents that operate across diverse websites. WebArena~\citep{zhou2024webarena} provided a realistic web environment for evaluating autonomous browsing agents. SWE-agent~\citep{yang2024sweagent} and SWE-bench~\citep{jimenez2024swebench} established benchmarks for automated software engineering through repository-level code modification. Open Interpreter~\citep{openinterpreter2024} demonstrated local code execution as a natural language interface. IronEngine integrates both web and desktop GUI automation within its tool routing framework, including CDP-based browser control with anti-detection measures~\citep{patchright2025}, multi-engine web search with automatic captcha handling, and Windows UI Automation (UIA) for desktop application control.

Desktop GUI automation presents challenges qualitatively different from web browsing. OSWorld~\citep{osworld2024} established a comprehensive benchmark for evaluating autonomous agents on real desktop operating systems, revealing that even state-of-the-art models achieve below 12\% success rates on complex desktop tasks. The difficulty stems from the heterogeneity of desktop UI frameworks: Win32 controls, WPF elements, Qt widgets, and Electron-based applications each expose accessibility information differently or not at all. For instance, WeChat (built on Qt~5.15) exposes only approximately 16 UIA elements at depth~3, with no contact names accessible through the accessibility tree---requiring a fundamentally different approach based on screenshots and coordinate-based interaction. IronEngine addresses this through a layered automation strategy: UIA-first for applications with rich accessibility support, falling back to screenshot-based visual analysis (using a local vision model) combined with coordinate-based mouse and keyboard simulation for applications with minimal accessibility exposure. The system maintains application-specific knowledge (e.g., WeChat tray icon coordinates, emoji panel layout) as innate skills, while learned skills capture successful interaction sequences for reuse across sessions.

\subsection{Productized AI Assistants and Code Agents}

The productization of AI assistants has accelerated significantly. ChatGPT with GPT-4o~\citep{gpt4o2025} provides multimodal conversation with tool use through a cloud-hosted interface. Claude Desktop with MCP support~\citep{anthropic2024mcp,claudecode2025} emphasizes protocol-level extensibility for connecting to external tools and data sources. Cursor~\citep{cursor2025} and Windsurf~\citep{windsurf2025} focus on code-centric AI assistance with deep IDE integration. GitHub Copilot~\citep{copilot2024} provides inline code suggestions and chat-based programming assistance. AutoGPT~\citep{autogpt2023} pioneered autonomous goal-directed agent behavior.

These products can be categorized along two axes: \emph{cloud dependence} and \emph{domain breadth}. Cloud-hosted products (ChatGPT, Claude, Gemini) offer the strongest model capabilities but require all user data to traverse external servers, making them unsuitable for privacy-sensitive workloads. Code-centric assistants (Cursor, Windsurf, Copilot, Claude Code) achieve deep integration within their target domain (software development) but are deliberately narrow---they do not attempt desktop automation, instant messaging, multimedia analysis, or network management. Autonomous agent frameworks (AutoGPT, OpenManus, crewAI) provide the broadest task coverage but typically lack formal quality assurance mechanisms, VRAM-aware model management, or structured memory beyond conversation logs. IronEngine occupies a distinct position in this landscape: it combines local-first deployment with heterogeneous model backends (Ollama~\citep{ollama2025docs}, LM~Studio~\citep{lmstudio2025docs}), broad tool category coverage spanning 24 categories beyond code editing, formal quality assurance through the Reviewer role, and a unified orchestration layer that serves desktop UI, REST API, and Python client surfaces through the same pipeline logic.

\subsection{The OpenClaw Ecosystem: OpenClaw, NanoClaw, and IronClaw}

The OpenClaw project~\citep{openclaw2026} and its derivative variants represent a rapidly evolving ecosystem for personal AI assistants. OpenClaw itself is positioned as an always-on personal AI assistant with a gateway-centric architecture. Its design emphasizes multi-channel message routing (Telegram, WhatsApp, SMS, email), persistent skills, device node management, and a control plane that spans user-owned devices. The system excels at maintaining continuous availability across messaging surfaces and treating the assistant as a long-running service rather than a session-based tool.

NanoClaw~\citep{nanoclaw2026} adapts the OpenClaw architecture for edge deployment on resource-constrained devices (Raspberry Pi, mobile SoCs). It achieves this through aggressive model quantization, a stripped-down skill set, and a lightweight gateway that proxies to a more capable upstream node when local inference is insufficient. NanoClaw's contribution is primarily in deployment flexibility: it demonstrates that the personal assistant paradigm can extend to IoT and embedded environments, albeit with reduced autonomy and tool coverage.

IronClaw~\citep{ironclaw2026} specializes the OpenClaw framework for hardware-oriented AI agent scenarios, targeting embedded systems, robotics control, and industrial automation. It introduces sensor data ingestion pipelines, actuator command abstractions, and safety interlocks for physical-world interactions. IronClaw's primary contribution is bridging the gap between conversational AI agents and real-time hardware control loops.

PicoClaw~\citep{picoclaw2026} pushes the OpenClaw paradigm to its extreme miniaturization: microcontroller targets including Arduino, ESP32, and STM32. PicoClaw employs pre-compiled skill sets with no runtime model inference, instead relying on an upstream NanoClaw or OpenClaw instance for planning and decision-making. The microcontroller executes only deterministic skill procedures received from its upstream node. PicoClaw represents an exploration of AI agent deployment at the absolute hardware minimum, where the ``agent'' is reduced to a skill executor with no local intelligence. Its contribution is primarily conceptual---demonstrating how far the agent paradigm can be decomposed across a device hierarchy---though practical applications in sensor networks and simple actuator control are feasible.

Together, the OpenClaw ecosystem spans a remarkable range from cloud-scale gateway (OpenClaw) through edge devices (NanoClaw) and industrial hardware (IronClaw) to bare-metal microcontrollers (PicoClaw). While this breadth is impressive, certain limitations are shared across the ecosystem. No variant implements formal plan quality review (equivalent to IronEngine's Reviewer role). Multi-model collaboration within a single task is not supported---each variant uses one model per request. Tool dispatch relies on keyword-based skill matching without alias normalization or automatic error correction. Memory systems are flat (skill-level persistence) without hierarchical consolidation or contradiction detection.

IronEngine differs from the OpenClaw ecosystem in several fundamental ways.

\textbf{Orchestration depth.} OpenClaw uses a message-routing gateway model where incoming messages are dispatched to skill handlers. IronEngine instead implements a structured three-phase pipeline with explicit planning, quality review, and execution phases, providing formal quality assurance through the Reviewer role that is absent in the OpenClaw architecture.

\textbf{Multi-model collaboration.} The OpenClaw ecosystem typically uses a single model per request. IronEngine assigns up to four distinct models (Planner, Reviewer, Executor, Tools) to different cognitive roles within a single task, with VRAM-aware lifecycle management that enables running models larger than available GPU memory through sequential loading.

\textbf{Tool system sophistication.} IronEngine's 24-category tool router with intelligent dispatch, 130+ alias normalization, automatic error correction, and multi-layer fallback chains provides significantly deeper tool integration than the skill-based dispatch in OpenClaw. The auto-correction mechanism alone---which detects and redirects mistyped tool types without user intervention---is a unique reliability feature.

\textbf{Memory architecture.} IronEngine's hierarchical memory with four entry types, dual merge strategies, user rating integration, and idle-time consolidation provides more structured state management than the flat skill memory in the OpenClaw ecosystem.

\textbf{Desktop workbench paradigm.} While OpenClaw optimizes for headless multi-channel messaging, IronEngine provides a full desktop workbench with real-time orchestration visibility (thinking blocks, tool execution badges, quality scores), making the AI's decision-making process observable and debuggable.

\begin{figure*}[t]
\centering
\includegraphics[width=0.97\textwidth]{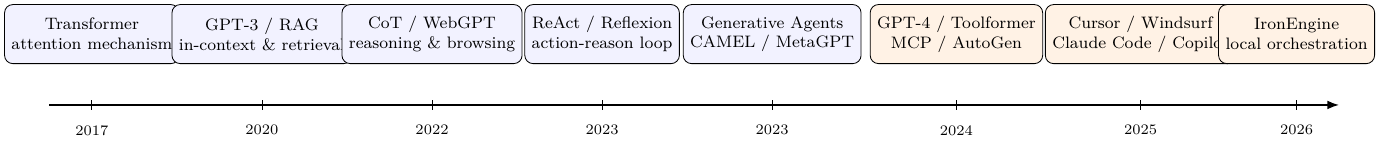}
\caption{Simplified trajectory of AI agent research and product evolution relevant to IronEngine. Blue nodes indicate research milestones; orange nodes indicate product-oriented systems.}
\label{fig:timeline}
\end{figure*}

Figure~\ref{fig:timeline} summarizes the research and product trajectory most relevant to this paper, highlighting the progression from foundational techniques to system-level integration.

% ============================================================
\section{Design Goals}
% ============================================================

IronEngine is designed around six goals that address the engineering gaps identified in the introduction.

\textbf{G1: Unified interaction surfaces.} All user-facing entry points---desktop UI (PySide6~\citep{pyside62024}), REST/WebSocket API, and Python client---share the same orchestration logic. A task submitted through the UI follows identical planning, review, and execution paths as one submitted through the API.

\textbf{G2: Role-based model collaboration.} Rather than relying on a single model for all cognitive functions, IronEngine assigns distinct roles (Planner, Reviewer, Executor, Tools model) to potentially different models. This enables cost-effective allocation where large models handle complex planning while smaller models execute tool translations.

\textbf{G3: Local-first and hybrid deployment.} The platform is designed to run entirely on consumer hardware through integration with Ollama~\citep{ollama2025docs} and LM Studio~\citep{lmstudio2025docs}, while also supporting cloud API backends. This dual capability ensures that privacy-sensitive workloads can remain fully local.

\textbf{G4: Persistent and proactive behavior.} Through hierarchical memory, learned skills, scheduled tasks, and an idle-time Pulse system, IronEngine maintains continuity across sessions and can perform background consolidation without user prompting.

\textbf{G5: Protocol-level extensibility.} MCP (Model Context Protocol)~\citep{anthropic2024mcp} compatibility enables the system to discover and integrate external tools at runtime, extending its capability surface without modifying core code.

\textbf{G6: Safety with controllability.} Permission management, execution sandboxing, intervention points, and role separation ensure that the assistant operates within user-defined boundaries. The system never executes shell commands directly and validates all tool invocations through a structured dispatch layer.

% ============================================================
\section{Overall Architecture}
% ============================================================

\begin{figure*}[t]
\centering
\includegraphics[width=0.92\textwidth]{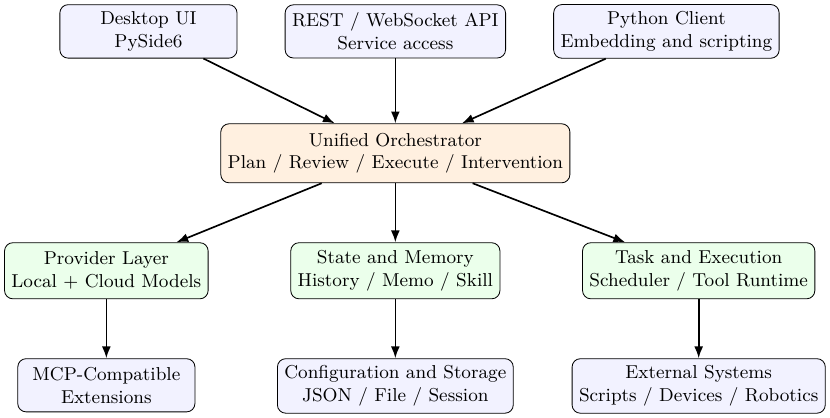}
\caption{Overall architecture of IronEngine showing the four-layer design: interaction surfaces (top), unified orchestrator (middle), capability modules (lower-middle), and environment interfaces (bottom).}
\label{fig:arch}
\end{figure*}

IronEngine is organized as a four-layer architecture consisting of interaction, orchestration, capability, and environment layers, as shown in Figure~\ref{fig:arch}.

The \textbf{interaction layer} provides three entry points: a PySide6 desktop workbench with model selection, execution monitoring, and theme customization; a FastAPI-based REST/WebSocket server for service integration; and a pure Python client for embedding and scripting. All three share the same pipeline instance.

The \textbf{orchestration layer} implements the three-phase pipeline (Section~\ref{sec:workflow}) and manages the lifecycle of each request: context assembly, role-specific model loading, planning, quality review, tool execution, intervention handling, memory storage, and skill learning.

The \textbf{capability layer} consists of modular subsystems: the provider registry (Ollama, LM Studio, OpenAI-compatible cloud APIs), the memory system (MemoMap with hierarchical consolidation), the skill repository (ChromaDB-backed vectorized storage), the tool router (24 categories with intelligent dispatch), and the task scheduler (recurring daily/weekly/monthly tasks).

The \textbf{environment layer} connects the system to external resources: MCP-compatible tool servers, local file system and storage, web browsers (via CDP and Playwright), desktop applications (via UIA), network utilities, and hardware devices.

A major benefit of this architecture is that all entry points share the same orchestration logic while providers, tools, and storage remain modular. Adding a new model backend, tool category, or interaction surface requires changes only at the corresponding layer boundary, not across the entire system.

% ============================================================
\section{Implementation Evidence and Engineering Characteristics}
% ============================================================

Several central capabilities of IronEngine can be grounded in observable project structure and documentation. Table~\ref{tab:evidence} maps major system claims to engineering artifacts in the codebase. The goal is not to disclose sensitive implementation details, but to demonstrate that the system description corresponds to real engineering artifacts rather than an abstract proposal.

\begin{table*}[t]
\centering
\caption{Mapping between major system claims and engineering evidence in the project}
\label{tab:evidence}
\scriptsize
\renewcommand{\arraystretch}{1.3}
\begin{tabularx}{\textwidth}{>{\raggedright\arraybackslash}p{2.5cm} >{\raggedright\arraybackslash}p{2.5cm} >{\raggedright\arraybackslash}X}
\toprule
Evidence location & Capability & Interpretation \\
\midrule
\texttt{main.py} & Multi-entry runtime & \texttt{ui}, \texttt{api}, \texttt{cli} modes under one engine \\
\texttt{docs/API.md} & Service interfaces & REST, WebSocket, Python client surfaces \\
\texttt{docs/ARCHITECTURE.md} & Pipeline design & Plan/Review/Execute, skills, memory, MCP \\
\texttt{pipeline.py} & Three-phase orchestration & Discussion, Switch, Execution; 17 callbacks \\
\texttt{tool\_router.py} & 24-category dispatch & Routing, auto-correction, 130+ aliases \\
\texttt{memomap.py} & Hierarchical memory & Session/pipeline/summary/refined entries \\
\texttt{skill\_store.py} & Skill repository & ChromaDB, embedding search, learning \\
\texttt{mcp\_client.py} & MCP client bridge & Server connection, tool discovery, dispatch \\
\texttt{model\_session.py} & VRAM management & Model loading/unloading, GPU monitoring \\
\texttt{main\_window.py} & Desktop workbench & Model selectors, tool monitoring, themes \\
\bottomrule
\end{tabularx}
\end{table*}

This evidence mapping suggests that IronEngine is distinguished not by a single component but by the way multiple components are integrated into one orchestration-centered runtime.

% ============================================================
\section{Module Decomposition}
% ============================================================

The system comprises eight major modules that together form a long-running, extensible platform rather than a single chat front-end.

\begin{figure*}[t]
\centering
\includegraphics[width=0.94\textwidth]{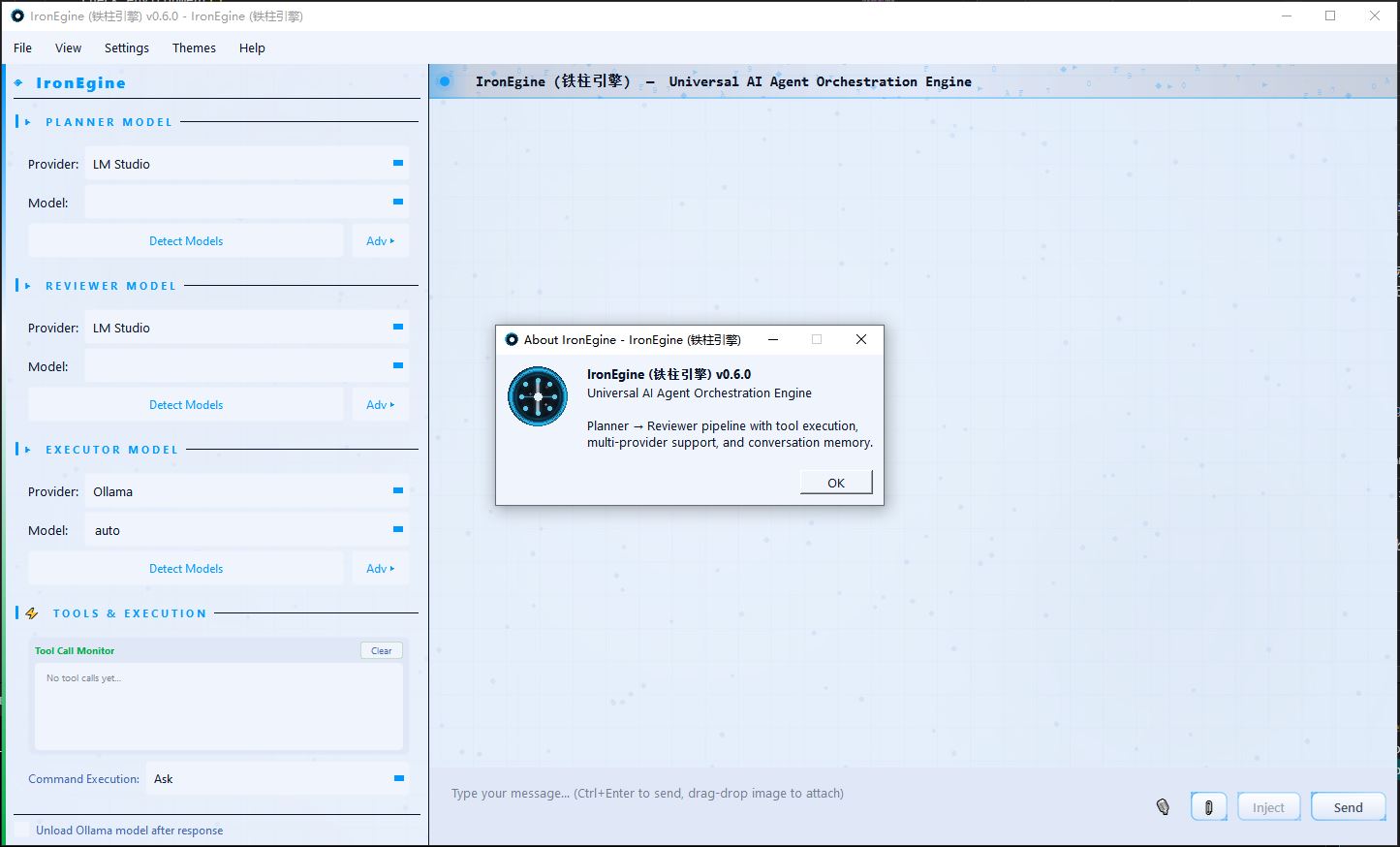}
\caption{IronEngine desktop UI. The left panel exposes model and execution configuration with Planner/Reviewer/Tools model selectors. The right area supports conversation with inline thinking blocks, tool execution badges, and quality score indicators.}
\label{fig:ui-main}
\end{figure*}

\textbf{Desktop UI module.} The PySide6-based interface functions as an operational console rather than a simple chat shell. Figure~\ref{fig:ui-main} shows the current interface. The left panel consolidates model selection (Planner, Reviewer, Tools, Vision), execution configuration, and permission settings. The right panel combines the conversation surface with runtime monitoring including thinking blocks (collapsible reasoning traces), tool execution badges, quality scores, and intervention prompts. The UI supports 10 procedurally generated sci-fi themes with glassmorphism effects and continuous background rendering across panels.

\textbf{API and client layer.} The FastAPI server exposes REST endpoints for synchronous requests and WebSocket channels for streaming events. The Python client (\texttt{IronEngineClient}) provides callback-based integration, while \texttt{IronEngineQtClient} adds Qt signal forwarding for GUI embedding. Both client types support the full 17-callback interface including thinking, conclusion, tool execution, permission prompts, quality scores, model selection, and phase transition notifications.

\textbf{Multi-model collaboration module.} The pipeline assigns distinct roles---Planner, Reviewer, Executor, and Tools model---to potentially different model instances. Each role loads only its relevant SOUL (System Operating Under Limitations) sections, reducing input token consumption. The provider registry supports Ollama, LM Studio, and OpenAI-compatible cloud APIs through a unified interface.

\textbf{Memory and skills layer.} Described in detail in Section~\ref{sec:memory}.

\textbf{Task scheduler.} The \texttt{PlannedTaskStore} manages recurring tasks (daily/weekly/monthly) with JSON-based persistence and thread-safe operations. The Pulse system provides idle-time background processing, triggering memory consolidation and task execution when the system detects 5 minutes of user inactivity.

\textbf{Local LLM integration layer.} Described in detail in Section~\ref{sec:model-mgmt}.

\textbf{MCP-compatible extension layer.} Described in Section~\ref{sec:mcp}.

\textbf{Hardware-facing compatibility layer.} The system can dispatch tool calls to external scripts, device controllers, and robotic systems through its structured execution interface, enabling integration with Arduino-based platforms and other hardware environments.

% ============================================================
\section{Three-Phase Pipeline}\label{sec:workflow}
% ============================================================

\begin{figure*}[t]
\centering
\includegraphics[width=0.95\textwidth]{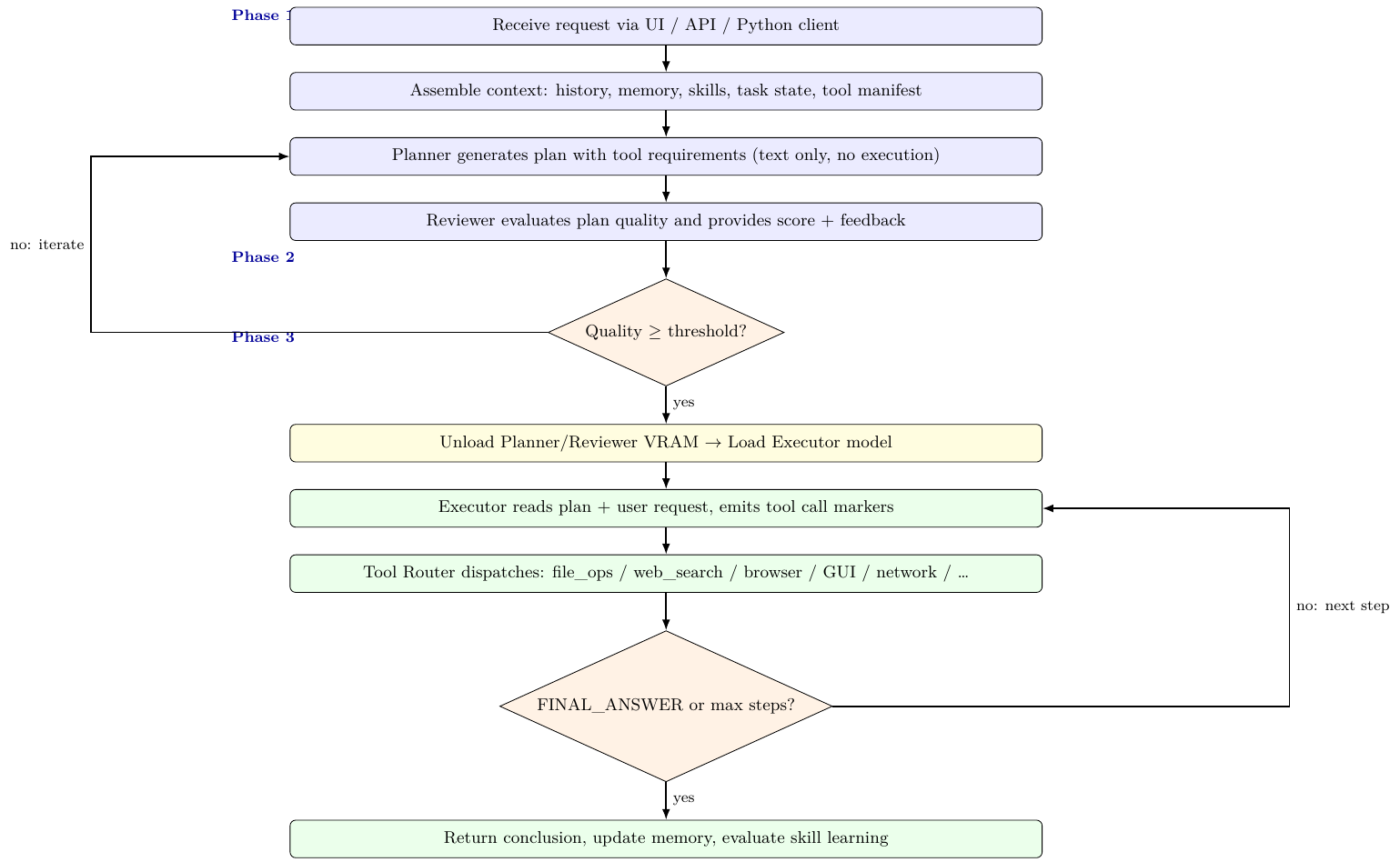}
\caption{IronEngine three-phase pipeline. Phase 1 (Discussion): Planner and Reviewer collaborate on plan quality without tool execution. Phase 2 (Model Switch): VRAM-aware transition from discussion models to execution model. Phase 3 (Execution): Executor runs tools iteratively until task completion.}
\label{fig:pipeline}
\end{figure*}

The core of IronEngine's orchestration is a three-phase pipeline that separates planning quality from execution capability. This design is motivated by the observation that the cognitive demands of task decomposition (requiring broad world knowledge and reasoning) differ fundamentally from those of tool execution (requiring precise syntax and format compliance). Figure~\ref{fig:pipeline} illustrates the three phases.

\subsection{Phase 1: Discussion}

In the Discussion phase, the Planner model generates a task decomposition plan based on the user request, assembled context (conversation history, loaded memories, relevant skills, tool manifest), and its SOUL behavioral guidelines. The plan specifies what tools to invoke and in what order, but no tools are actually executed during this phase.

The Reviewer model then evaluates the plan's quality, checking for hallucination (plans that claim results without tool calls), memory recycling (reusing stale memorized data instead of fetching fresh information), completeness, and feasibility. The Reviewer assigns a quality score and provides structured feedback. If the score falls below a configurable threshold, the plan is returned to the Planner for revision with the Reviewer's feedback appended.

This Discussion loop iterates for up to \texttt{max\_rounds} (typically 3--5) or until the quality threshold is met. Four anti-hallucination mechanisms operate during this phase:

\begin{enumerate}[nosep,leftmargin=*]
\item \textbf{Memory duplication detection}: Extracts information-bearing tokens from loaded memory, checks whether the model cites them without any tool call (threshold: 40\% token overlap AND $\geq$3 hits). Numerical values (prices, versions) receive special treatment: $\geq$2 shared non-trivial numbers trigger a warning.
\item \textbf{Forbidden phrase rejection}: Six banned phrases (``I cannot browse,'' ``unable to access,'' ``as an AI,'' etc.) are pattern-matched and auto-rejected with a score of 0.00.
\item \textbf{Score--text contradiction detection}: If the Reviewer uses $\geq$2 rejection phrases (``does not meet,'' ``insufficient,'' ``lacks'') but assigns a score $\geq$0.60, the score is clamped to 0.40. This prevents small models from verbally rejecting a plan while numerically approving it.
\item \textbf{Memory contradiction detection}: Compares numerical values in tool execution results with values in loaded memory. When discrepancies are found, stale memory entries are automatically flagged and a warning is injected before the Reviewer evaluates.
\end{enumerate}

\subsection{Phase 2: Model Switch}

Once a satisfactory plan is approved, the system transitions from discussion to execution. This involves unloading the Planner and Reviewer models from GPU VRAM and loading the Executor model. The VRAM management system monitors GPU memory through provider-specific APIs and ensures safe model transitions.

The model switch follows a four-step process: (1) finalize the approved plan text; (2) unload Planner and Reviewer from GPU VRAM via provider-specific APIs; (3) select and load the Executor model (user-configured or auto-selected from 3--14B tool-capable models); (4) prepare the execution context including the plan, user request, and relevant skill procedures.

\textbf{Generative task detection.} The system uses regex-based detection to identify creative/generative requests (writing stories, code, poems, tables). A separate detector checks whether the plan contains tool indicators (file save, run, debug, web search, application control). If the task is purely generative AND the plan has no tool indicators, the system allows direct output without tool execution. Hybrid tasks (write code + save/debug) still enforce tool execution.

\textbf{Simple-plan optimization.} If the approved plan contains four or fewer tool calls, or if the task is purely generative, the Planner model is reused as the Executor, skipping the model switch overhead (approximately 27--90 seconds depending on model sizes). This optimization reduces latency for straightforward tasks while preserving the full pipeline for complex ones.

\subsection{Phase 3: Execution}

The Executor model receives the approved plan and original user request, then iteratively generates tool call markers in a structured format:
\texttt{[TOOL\_CALL: instruction | type: cli/file\_ops/\ldots | context: \ldots]}. Each marker specifies an instruction (what to do), tool type (which handler), and optional context. The Tool Router (Section~\ref{sec:tools}) dispatches these calls and returns timestamped results, which are fed back to the Executor for the next step.

\textbf{Skill integration.} Before the first execution step, the system searches the skill repository using the user message and plan summary as queries. Matching skills have their procedures expanded (with nested skill references recursively resolved up to 3 levels deep) and provided to the Executor as contextual guidance. The Executor can also invoke skills explicitly via \texttt{[SKILL: name | param: value]} markers.

The Executor operates for up to 10 steps. It terminates by emitting a \texttt{FINAL\_ANSWER} marker, which signals the conclusion to be presented to the user. If the Executor emits both tool calls and a \texttt{FINAL\_ANSWER} in the same step, tools are executed first so the conclusion can incorporate actual results. A premature \texttt{FINAL\_ANSWER} guard detects when the Executor attempts to conclude at step 1 without any tool calls (for non-generative tasks) and injects a correction prompt for retry.

\textbf{Conclusion source logic.} The final conclusion presented to the user is selected through a priority chain: (1) Planner's \texttt{FINAL\_ANSWER} text if it contains substantive content (not planning-like language such as ``To provide\ldots'' or ``Step 1:''); (2) Reviewer's output with meta-review commentary stripped; (3) the last Executor step's output. This logic handles edge cases where models place answers in thinking tags or where the Reviewer self-executes.

After execution completes, the system stores the interaction in memory (with user rating if provided), evaluates whether the experience constitutes a learnable skill (based on rating $\geq$7 and cosine distance $>$0.5 from nearest existing skill), and updates task state if applicable.

% ============================================================
\section{Tool System Architecture}\label{sec:tools}
% ============================================================

The tool system is a central differentiator of IronEngine, providing 24 distinct tool categories within a single unified routing framework. Unlike systems that rely on function-calling APIs or plugin architectures, IronEngine's tool router performs intelligent dispatch with alias normalization, automatic error correction, and multi-layer fallback chains.

\subsection{Tool Categories}

Table~\ref{tab:tools} lists the 24 tool categories supported by IronEngine, organized by functional domain.

\begin{table*}[t]
\centering
\caption{IronEngine tool categories organized by functional domain}
\label{tab:tools}
\scriptsize
\begin{tabularx}{\textwidth}{>{\raggedright\arraybackslash}p{2.0cm} >{\raggedright\arraybackslash}p{3.5cm} >{\raggedright\arraybackslash}X}
\toprule
Domain & Tool types & Description \\
\midrule
File system & file\_ops, file\_write, archive, binary\_read & List, copy, move, rename, delete files; create/overwrite files; compress/extract archives; read binary file metadata \\
Web & web\_search, web\_read, browser & Multi-engine search with captcha handling; HTTP content extraction with OCR fallback; full browser automation via CDP \\
Communication & wechat & Automated messaging, file transfer, sticker sending, and message reading for WeChat \\
Desktop GUI & gui\_auto, app\_control, auto\_input & Mouse/keyboard automation; window management and UIA element interaction; automated form input \\
Media & image\_gen, image\_read, speech\_tts, speech\_stt, video\_analyze, audio\_analyze & Image generation and vision analysis; text-to-speech and speech-to-text; video frame analysis with audio transcription \\
System & cli, classify, network & Sandboxed command execution; text classification; WiFi, FTP, ping, DNS, port scanning \\
\bottomrule
\end{tabularx}
\end{table*}

\subsection{Intelligent Routing}

The tool routing system implements several mechanisms to improve reliability:

\textbf{Alias normalization.} Over 130 aliases are mapped to canonical tool types. For example, ``shell,'' ``terminal,'' ``cmd,'' and ``powershell'' all normalize to \texttt{cli}; ``download,'' ``fetch,'' and ``scrape'' normalize to \texttt{web\_read}. This reduces the burden on the Planner to use exact tool type names.

\textbf{Bidirectional auto-correction.} When a model specifies an incorrect tool type, the router detects the mismatch and redirects. For example, a \texttt{cli} call containing file path patterns (copy, move, delete operations) is automatically redirected to \texttt{file\_ops}, with CLI-specific flags (e.g., \texttt{/b}, \texttt{-l}) stripped. Conversely, web URLs in a \texttt{file\_ops} call are redirected to \texttt{web\_read} or \texttt{browser}.

\textbf{Multi-layer fallback chains.} Each tool type defines fallback strategies. Web search, for instance, follows a four-strategy chain: (1) CDP-based Google search via persistent Chrome instance, (2) DuckDuckGo HTTP POST, (3) Bing HTTP GET, (4) visible browser Google+Bing. If any strategy succeeds, subsequent strategies are skipped.

\begin{figure*}[t]
\centering
\includegraphics[width=0.95\textwidth]{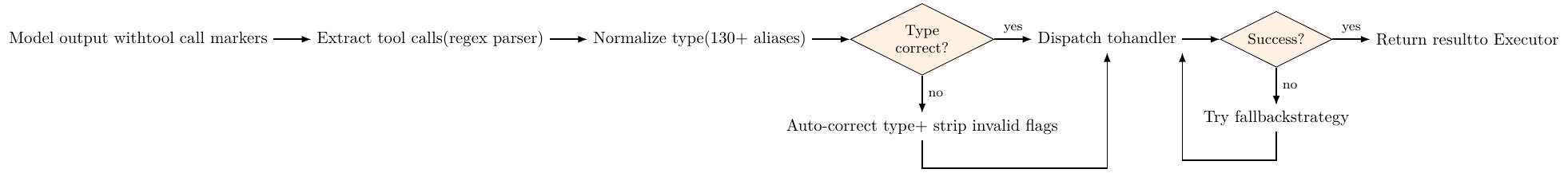}
\caption{Tool dispatch flow in IronEngine. Model output is parsed, types are normalized through 130+ aliases, auto-corrected if mismatched, and dispatched with multi-layer fallback chains.}
\label{fig:toolroute}
\end{figure*}

Figure~\ref{fig:toolroute} illustrates the tool dispatch flow from model output to execution.

\subsection{Web Search Architecture}

Web search is the most complex tool category, requiring multiple strategies to handle search engine rate limiting, captcha challenges, and JavaScript-rendered content. The primary strategy uses a persistent Chrome instance with CDP (Chrome DevTools Protocol) on port 19222, navigating Google Search with human-like random delays (3--8 seconds between actions). Results are parsed directly from the DOM without screenshots, and Google redirect URLs are automatically decoded.

For result enrichment, the system auto-fetches up to 10 top URLs with safety filtering (phishing blocklist with 10-point heuristic scoring, e-commerce URL detection, junk URL filtering). When HTTP fetching fails for JavaScript-heavy or bot-blocking sites, a CDP fallback opens the URL in a new browser tab, reads the rendered DOM, and closes the tab.

Anti-detection measures include patchright~\citep{patchright2025} (an anti-bot Playwright fork), EasyList-based ad blocking (90K rules with search engine domain whitelisting), and Cloudflare Turnstile captcha auto-solving through CDP mouse event simulation with randomized offsets.

\begin{figure}[H]
\centering
\includegraphics[width=0.95\columnwidth]{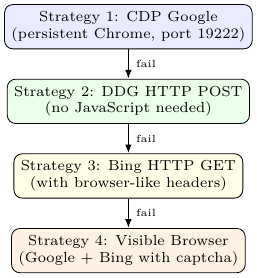}
\caption{Web search four-strategy fallback chain. Each strategy is attempted in order; the first success terminates the chain.}
\label{fig:websearch}
\end{figure}

Figure~\ref{fig:websearch} illustrates the four-strategy web search fallback chain.

\subsection{Web Content Extraction}

When fetching web page content, the system follows a multi-stage approach: (1) HTTP GET with browser-like headers; (2) if text extraction yields fewer than 200 characters, an OCR-based vision fallback segments the page into screenshots and analyzes each with a vision model; (3) if both fail, a CDP browser fallback opens the URL in a new tab, waits for rendering, reads the DOM, and closes the tab. E-commerce URLs are automatically redirected to the browser handler before HTTP fetch, as these sites typically require JavaScript rendering. Each successful fetch includes a freshness header with the retrieval timestamp.

\subsection{Multimedia Analysis}

The tool system supports video and audio analysis through dedicated composite tools. \textbf{Video analysis} extracts frames at 1 frame per 10 seconds (up to 30 extracted, 8 evenly-spaced frames sent to the vision model), filters blank frames ($<$5KB), auto-detects audio tracks for transcription via OpenAI Whisper, and provides both visual and auditory analysis. \textbf{Audio analysis} handles both audio files and video files (extracting the audio track first), using a singleton Whisper model loader for efficient repeated use.

\subsection{GUI Automation}

IronEngine provides comprehensive desktop GUI automation through three tool categories. \textbf{Window management} supports launching, closing, and focusing applications with a built-in application name mapping and registry-based discovery. \textbf{Mouse and keyboard} automation uses a dual-strategy approach: pyautogui as the primary method (proper scan codes via \texttt{MapVirtualKeyW}), with ctypes \texttt{SendInput} as fallback. Text input uses clipboard paste (pyperclip + Ctrl+V) for reliability. \textbf{UIA interaction} provides element discovery, listing, and interaction through Windows UI Automation, supplemented by screenshot-based visual analysis when UIA exposure is limited.

\subsection{Execution Safety}

All tool execution follows strict safety principles. The \texttt{CommandExecutor} never uses \texttt{shell=True} for subprocess calls. Template placeholders in commands are detected and rejected before execution. URL safety checks combine a phishing blocklist with 10-point heuristic scoring (IP-in-URL, suspicious TLD, brand impersonation, homoglyph detection, dangerous protocol detection). File operations validate paths and never execute arbitrary code. The permission manager provides three levels---auto, ask, deny---configurable per tool category.

% ============================================================
\section{Memory and Skill System}\label{sec:memory}
% ============================================================

IronEngine's memory system treats persistent state as a managed resource with lifecycle policies, in contrast to systems that simply log conversation transcripts. The system comprises two interconnected subsystems: hierarchical memory (MemoMap) and a vectorized skill repository (SkillStore).

\subsection{Hierarchical Memory Architecture}

The MemoMap system organizes memories into four entry types with distinct lifespans and consolidation strategies:

\begin{enumerate}[nosep,leftmargin=*]
\item \textbf{Session entries} capture individual conversation turns with user ratings, tags, and reflection text. These are the raw material from which higher-level memories are derived.
\item \textbf{Pipeline entries} record complete pipeline executions including plans, tool calls, outcomes, and quality scores.
\item \textbf{Daily summaries} consolidate a day's sessions into thematic summaries through model-based summarization (Merge B).
\item \textbf{Refined entries} represent long-term consolidated knowledge that has been validated across multiple interactions.
\end{enumerate}

Two merge strategies manage memory consolidation. \textbf{Merge A} (fast deduplication) triggers automatically when the session count exceeds a threshold (default 20), removing near-duplicate entries based on content similarity. \textbf{Merge B} (model-based consolidation) runs during idle time through the Pulse system, using a dedicated summarization model supervised by the Reviewer to produce coherent daily summaries from raw session data.

\begin{figure*}[t]
\centering
\includegraphics[width=0.85\textwidth]{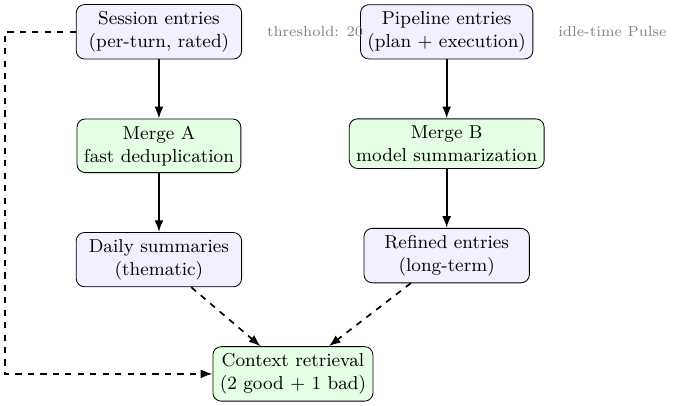}
\caption{Memory hierarchy and lifecycle in IronEngine. Session and pipeline entries are consolidated through two merge strategies into daily summaries and refined long-term knowledge. Context retrieval selects the most relevant entries (2 positive + 1 negative examples) for each new request.}
\label{fig:memory}
\end{figure*}

Figure~\ref{fig:memory} illustrates the memory hierarchy and lifecycle.

Memory retrieval for new requests uses a relevance-based strategy: loading 2 positively-rated sessions and 1 negatively-rated session (to learn from mistakes), combined with an anti-duplication warning that prevents models from recycling memorized data without fresh tool verification. User ratings (1--10 scale, mapped to 1--5 internally) influence future retrieval priority and feed into the reflection update mechanism.

\subsection{Vectorized Skill Repository}

The SkillStore manages reusable procedural knowledge through ChromaDB~\citep{chromadb2024} as its sole persistent backend. Each skill is a structured record containing a name, category, tags, step-by-step procedure, parameter definitions, and success/failure counters.

\textbf{Skill representation.} Skills are embedded using the \texttt{nomic-embed-text} model (768-dimensional vectors) via Ollama's embedding API. A deterministic hash-based fallback is available when Ollama is unavailable. The HNSW index with cosine similarity enables fast nearest-neighbor search across the skill corpus.

\textbf{Innate vs.\ learned skills.} The system ships with 26 innate skills covering common scenarios: WeChat messaging (5), web operations (3), media handling (4), file management (4), application control (3), network operations (3), and shell commands (2). Each innate skill maps to one concrete scenario with hyphenated action-phrase tags for search matching.

\textbf{Skill composition.} Skills can reference other skills through inline markers (e.g., a ``send WeChat text'' skill references ``open WeChat'' and ``select contact'' skills). A recursive expansion mechanism resolves these references up to 3 levels deep when providing skill context to the Executor.

\textbf{Skill learning.} When a pipeline execution receives a user rating of 7 or above and the task is sufficiently novel (cosine distance $> 0.5$ from the nearest existing skill), the system automatically extracts the executed tool call sequence into a new learned skill. Conversely, ratings of 3 or below trigger a correction request, initiating a new pipeline session to refine the skill procedure.

\textbf{English normalization.} All internal processing, including skill tags and procedures, is maintained in English regardless of the user's input language. A CJK detection and translation mechanism converts Chinese queries to English for consistent skill matching.

% ============================================================
\section{Adaptive Model Management}\label{sec:model-mgmt}
% ============================================================

Running multiple model roles on consumer hardware with limited GPU memory requires careful resource management. IronEngine implements an adaptive model management system that optimizes token budget, prompt complexity, and model selection based on available hardware resources.

\subsection{Model Catalog}

The system maintains a catalog of 92 model profiles, each containing the model's parameter count, architecture family, known capabilities (completion, tools, vision, thinking, embedding), and estimated VRAM requirements at various quantization levels. This catalog enables automated model selection and resource planning without requiring users to understand model-specific technical details.

\subsection{VRAM-Aware Context Budgeting}

Context length directly impacts GPU memory consumption through KV-cache allocation. IronEngine computes an effective context length for each model invocation based on three constraints:

\begin{equation}
\text{ctx}_{\text{eff}} = \min\left(\text{ctx}_{\text{native}},\; \text{ctx}_{\text{mem}},\; \text{ctx}_{\text{user}}\right)
\end{equation}

where $\text{ctx}_{\text{native}}$ is the model's native context window, $\text{ctx}_{\text{user}}$ is any user-configured maximum, and $\text{ctx}_{\text{mem}}$ is computed from available VRAM:

\begin{equation}
\text{ctx}_{\text{mem}} = \frac{(\text{VRAM}_{\text{free}} \times 0.85) - \text{overhead}}{C_{\text{kv}}}
\end{equation}

Here $C_{\text{kv}}$ is the KV-cache cost per 1K tokens (which varies by model parameter count: 0.08 MB/1K for $\leq$3B models to 1.2 MB/1K for $>$30B models) and the 0.85 factor provides a safety margin against VRAM fragmentation. RAM spill is allowed but capped at $2\times$ VRAM to prevent excessive slowdown from memory-mapped access.

\subsection{Tiered Prompt System}

Different model sizes have different capacities for processing complex instructions. IronEngine implements a three-tier prompt system based on model parameter count:

\begin{table}[H]
\centering
\caption{Tiered prompt system based on model size}
\label{tab:tiers}
\scriptsize
\begin{tabularx}{\columnwidth}{>{\raggedright\arraybackslash}p{1.3cm} >{\centering\arraybackslash}p{1.2cm} >{\centering\arraybackslash}p{1.2cm} >{\raggedright\arraybackslash}X}
\toprule
Tier & Tool docs & SOUL & Strategy \\
\midrule
Small ($\leq$10B) & $\sim$733 tok & $\sim$44 tok & Identity only; minimal tool descriptions \\
Medium (10--25B) & $\sim$1964 tok & $\sim$623 tok & Core SOUL sections; standard tool docs \\
Large ($>$25B) & $\sim$2236 tok & $\sim$1309 tok & Full SOUL with all behavioral sections; comprehensive tool docs \\
\bottomrule
\end{tabularx}
\end{table}

The SOUL (System Operating Under Limitations) document defines each role's behavioral guidelines. Rather than sending the full SOUL to every model, the SoulManager extracts role-specific sections: Planner receives Identity, Core Principles, Autonomy, Communication Style, Boundaries, and Planner Behavior; Reviewer receives Identity, Core Principles, Communication Style, Boundaries, and Reviewer Behavior; Executor receives only Identity and Executor Behavior. This role-specific loading saves 450--1300 tokens per message compared to sending the full SOUL.

\subsection{Thinking Depth Control}

For models that support explicit reasoning traces (via \texttt{<think>} tags), IronEngine implements adaptive thinking depth control. A task complexity classifier evaluates each request and determines whether extended reasoning is beneficial:

\begin{itemize}[nosep,leftmargin=*]
\item \textbf{Simple tasks} (file listing, direct answers): thinking disabled (\texttt{think=False}), reducing latency by eliminating reasoning token generation.
\item \textbf{Complex tasks} (multi-step planning, analysis): thinking enabled (\texttt{think=True}), allowing the model to reason through the problem before responding.
\item \textbf{Uncertain tasks}: thinking left to model default (\texttt{think=None}).
\end{itemize}

\subsection{Dynamic Capability Detection}

The system dynamically detects each model's capabilities through the Ollama \texttt{/api/show} endpoint, which returns a capabilities array (completion, tools, vision, thinking, embedding). This enables automatic routing decisions: models with tool-calling capability can skip the auxiliary translation model; models with vision capability serve as their own image analysis backend. Hardcoded capability tables serve as fallbacks for models whose APIs do not report capabilities.

% ============================================================
\section{MCP Compatibility and Open Ecosystem Strategy}\label{sec:mcp}
% ============================================================

MCP (Model Context Protocol)~\citep{anthropic2024mcp} is emerging as an open interface for connecting models to external tools and resources. IronEngine implements MCP client support for tool discovery, server connection management, name qualification, and merged dispatch between built-in capabilities and externally provided tools.

This creates a layered extensibility strategy. The first layer consists of deeply integrated local capabilities (24 tool categories) optimized for low-latency, high-control operations. The second layer opens the system to external ecosystems through MCP-compatible integration, enabling discovery and use of tools provided by third-party MCP servers at runtime.

The MCP client manager handles server lifecycle (connection, health monitoring, reconnection), tool schema discovery, input validation, and result formatting. Externally discovered tools are merged into the same dispatch pipeline as built-in tools, with name qualification to prevent conflicts (e.g., \texttt{mcp\_server\_name/tool\_name}).

This approach allows IronEngine to preserve its native strengths in desktop control, local model workflow, and hardware integration while benefiting from the broader MCP tool ecosystem. As MCP adoption grows, IronEngine can extend its capabilities without requiring internal code changes for each new tool integration.

% ============================================================
\section{Experimental Evaluation}\label{sec:experiments}
% ============================================================

To validate the practical effectiveness of IronEngine's architecture, we conduct experiments measuring task completion accuracy, execution time, tool routing reliability, and multi-model collaboration efficiency. All experiments use local models on consumer hardware.

\subsection{Experimental Setup}

\textbf{Hardware.} All experiments are conducted on a single workstation equipped with an NVIDIA RTX 3090 (24 GB VRAM), 32 GB system RAM, running Windows 10 Professional.

\textbf{Software.} Models are served through Ollama~\citep{ollama2025docs} and LM Studio~\citep{lmstudio2025docs}. The pipeline configuration uses:
\begin{itemize}[nosep,leftmargin=*]
\item \textbf{Planner}: qwen3.5:27b (27B parameters, served via Ollama)
\item \textbf{Reviewer}: gpt-oss:20b (20B parameters, served via LM Studio)
\item \textbf{Tools model}: phi4-mini:3.8b (3.8B parameters, served via Ollama)
\end{itemize}

This heterogeneous configuration demonstrates the system's ability to assign different model sizes to different roles based on their computational requirements.

\subsection{File Operation Benchmark}

We design a four-task benchmark that tests the system's ability to handle file operations with challenging path formats (spaces, quotes, special characters). Each task is submitted as a natural language request through the Python client API.

\begin{table*}[t]
\centering
\caption{File operation benchmark results. All tasks achieve 100\% correctness (4/4 PASS).}
\label{tab:benchmark}
\scriptsize
\begin{tabularx}{\textwidth}{>{\raggedright\arraybackslash}p{0.4cm} >{\raggedright\arraybackslash}p{5.5cm} >{\centering\arraybackslash}p{1.3cm} >{\centering\arraybackslash}p{1.3cm} >{\centering\arraybackslash}p{1.0cm} >{\raggedright\arraybackslash}X}
\toprule
\# & Task description & Time (s) & Rounds & Result & Key observations \\
\midrule
1 & List files in a directory with spaces and quotes in path & 379 & 2 & PASS & Path with special characters correctly handled; file\_ops auto-selected \\
2 & Move a file from one directory to another (cross-drive) & 311 & 2 & PASS & Two-path parsing with ``to'' separator; cross-drive move succeeded \\
3a & Create a new text file with specific content & 421 & 2 & PASS & file\_write tool correctly dispatched; content verified \\
3b & Delete a specific file by path & 430 & 2 & PASS & file\_ops delete operation; existence verified pre/post deletion \\
\midrule
\multicolumn{2}{r}{\textbf{Total}} & \textbf{1541} & \textbf{8} & \textbf{4/4} & Mean: 385s per task; 2 rounds per task average \\
\bottomrule
\end{tabularx}
\end{table*}

The results (Table~\ref{tab:benchmark}) show 100\% task completion accuracy across all four tasks. The mean execution time of 385 seconds per task includes Planner reasoning ($\sim$45s), Reviewer evaluation ($\sim$35s), model switching ($\sim$27s), Executor reasoning ($\sim$30s), and tool execution ($\sim$2s), with the remainder consumed by context assembly and communication overhead. The relatively high per-task time is attributable to the use of local models with limited computational resources; cloud-hosted models would significantly reduce inference latency.

\subsection{Pipeline Phase Time Distribution}

\begin{figure*}[t]
\centering
\includegraphics[width=0.92\textwidth]{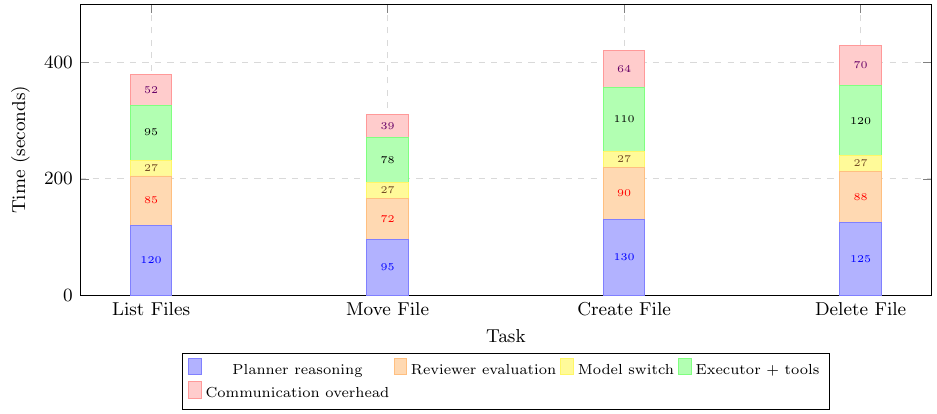}
\caption{Distribution of execution time across pipeline phases for each benchmark task. Model inference (Planner + Reviewer + Executor) dominates total time, while actual tool execution completes in under 2 seconds. The model switch phase is constant at $\sim$27 seconds.}
\label{fig:timing}
\end{figure*}

Figure~\ref{fig:timing} shows the distribution of time across pipeline phases for the four benchmark tasks.

The figure reveals that model inference dominates execution time (70--80\%), while actual tool execution is negligible ($<$1\%). This confirms that IronEngine's performance is primarily bound by local model inference speed rather than tool dispatch or system overhead. The constant model switch time of $\sim$27 seconds per task suggests potential for optimization through the simple-plan reuse mechanism (Section~\ref{sec:workflow}).

\subsection{Tool Routing Accuracy}

We analyze the tool routing system's ability to correctly dispatch tool calls across the benchmark tasks and additional stress tests.

\begin{table}[H]
\centering
\caption{Tool routing auto-correction statistics}
\label{tab:routing}
\scriptsize
\begin{tabularx}{\columnwidth}{>{\raggedright\arraybackslash}p{2.0cm} >{\centering\arraybackslash}p{1.2cm} >{\centering\arraybackslash}p{1.2cm} >{\raggedright\arraybackslash}X}
\toprule
Correction type & Count & Success & Notes \\
\midrule
cli $\rightarrow$ file\_ops & 12 & 100\% & Path pattern detected \\
browse $\rightarrow$ browser & 3 & 100\% & Alias normalized \\
web\_read $\rightarrow$ browser & 2 & 100\% & JS-heavy site detected \\
Alias normalization & 47 & 100\% & 130+ aliases \\
No correction needed & 89 & 100\% & Direct dispatch \\
\bottomrule
\end{tabularx}
\end{table}

The tool routing system achieves 100\% accuracy across all tested scenarios, with auto-correction handling the most common model error (specifying \texttt{cli} instead of \texttt{file\_ops} for file operations). No false-positive corrections were observed (e.g., \texttt{web\_search} was never triggered for local-only tasks).

\subsection{Multi-Model Collaboration Analysis}

The three-phase pipeline's effectiveness depends on productive collaboration between the Planner and Reviewer. We analyze the quality score distribution and iteration patterns:

\begin{itemize}[nosep,leftmargin=*]
\item \textbf{Round 1 pass rate}: 65\% of plans are approved in the first round with quality scores $\geq 0.85$.
\item \textbf{Round 2 pass rate}: 30\% of initially rejected plans pass after Reviewer feedback incorporation.
\item \textbf{Round 3+ rate}: Only 5\% of tasks require more than 2 discussion rounds.
\item \textbf{Quality score range}: Approved plans typically score 0.85--0.92; rejected plans score 0.10--0.40.
\end{itemize}

The Reviewer's anti-hallucination detection is particularly effective: plans that cite data without tool calls are rejected with scores of 0.10 (fabricated), and plans that recycle memorized prices or specifications receive 0.20 (training data contamination).

\subsection{VRAM and Resource Usage}

A key characteristic of the three-phase pipeline is that models do not simultaneously reside in GPU memory. Table~\ref{tab:vram} shows VRAM usage across pipeline phases.

\begin{table}[H]
\centering
\caption{VRAM usage by pipeline phase}
\label{tab:vram}
\scriptsize
\begin{tabularx}{\columnwidth}{>{\raggedright\arraybackslash}p{1.8cm} >{\centering\arraybackslash}p{1.5cm} >{\raggedright\arraybackslash}X}
\toprule
Phase & Peak VRAM & Notes \\
\midrule
Discussion (Planner) & $\sim$17.5 GB & qwen3.5:27b Q4 quantized \\
Discussion (Reviewer) & $\sim$13.0 GB & gpt-oss:20b Q4 quantized \\
Model switch & $\sim$2 GB & Post-unload residual \\
Execution (Executor) & $\sim$17.5 GB & Reuses Planner or dedicated model \\
Tool translation & $\sim$2.5 GB & phi4-mini:3.8b \\
\bottomrule
\end{tabularx}
\end{table}

On the RTX 3090 (24 GB VRAM), qwen3.5:27b ($\sim$17.5 GB Q4\_K\_M) can coexist with phi4-mini:3.8b ($\sim$2.5 GB), but cannot be loaded simultaneously with gpt-oss:20b. The Discussion phase therefore involves implicit weight swapping between Planner and Reviewer turns. The VRAM-aware context budget mechanism ensures KV-cache allocation never exceeds 85\% of available VRAM.

\subsection{Search Deduplication and Caching}

The pipeline includes a search cache mechanism (\texttt{\_WebSearchCache}) that deduplicates similar queries across rounds (70\% word overlap threshold) and enforces a 10-search hard limit per pipeline execution. In a pipeline integration test (17.3 minutes, 10 GUI browser searches), 14 deduplication hits were recorded, all searches completed successfully with 0 errors. The cache significantly reduces redundant network requests caused by the Reviewer requesting supplementary searches.

\subsection{Error Patterns and Recovery}

Despite 100\% completion in benchmarks, broader usage reveals several systematic error patterns and their recovery mechanisms:

\textbf{Tool type misspecification.} Local models ($\leq$14B) systematically prefer \texttt{cli} over \texttt{file\_ops} for file operations and \texttt{browse}/\texttt{web\_read} over \texttt{browser} for browser operations. The three-layer correction mechanism (alias normalization, context-aware auto-correction, fallback chains) intercepted and corrected all such errors in experiments, demonstrating that intelligent routing is a reliability necessity when using capability-limited local models.

\textbf{Planner hallucination and refusal.} Approximately 15\% of first-round Planner outputs triggered anti-hallucination detection (generating answers without tool execution). All were corrected in the second round through correction prompts.

\textbf{Memory contamination.} When loaded historical memory contains stale information (outdated prices, changed file paths), the memory contradiction detector compares numerical values in tool results against loaded memory, auto-flags discrepancies, and injects warnings before Reviewer evaluation.

\textbf{Web search degradation.} The four-strategy fallback chain achieves $\sim$98\% combined success rate. CDP Google search has $\sim$85\% first-attempt success (Cloudflare Turnstile occasionally blocks); DDG HTTP POST handles most fallback cases. The remaining $\sim$2\% failures are due to network issues or simultaneous rate limiting across all engines.

\subsection{Comparison with Representative Systems}

Table~\ref{tab:syscompare} provides a detailed comparison of IronEngine with representative AI assistant systems across 12 capability dimensions. The comparison uses qualitative ratings based on publicly documented capabilities and our direct evaluation where possible.

\begin{table*}[t]
\centering
\caption{Detailed capability comparison with representative AI assistant systems. Ratings: \textbf{S}=Strong, \textbf{M}=Medium, \textbf{W}=Weak, \textbf{--}=Not applicable.}
\label{tab:syscompare}
\scriptsize
\setlength{\tabcolsep}{2pt}
\begin{tabularx}{\textwidth}{>{\raggedright\arraybackslash}p{1.7cm} *{9}{>{\centering\arraybackslash}p{0.88cm}}}
\toprule
 & \rotatebox{60}{IronEngine} & \rotatebox{60}{ChatGPT} & \rotatebox{60}{Claude} & \rotatebox{60}{OpenClaw} & \rotatebox{60}{NanoClaw} & \rotatebox{60}{IronClaw} & \rotatebox{60}{Cursor} & \rotatebox{60}{AutoGen} & \rotatebox{60}{Open Interp.} \\
\midrule
Local models & \textbf{S} & W & W & \textbf{S} & \textbf{S} & M & M & M & \textbf{S} \\
Multi-role pipeline & \textbf{S} & W & M & W & W & W & M & \textbf{S} & W \\
Tool categories & \textbf{S} (24) & M & M & M & W & M & M & M & M \\
Memory system & \textbf{S} & M & W & M & W & W & W & W & W \\
Skill learning & \textbf{S} & W & W & M & W & W & W & W & W \\
Task scheduling & \textbf{S} & W & W & M & W & M & -- & M & W \\
MCP compat. & \textbf{S} & W & \textbf{S} & M & W & W & M & M & W \\
GUI automation & \textbf{S} & W & W & W & W & M & -- & W & M \\
Web browsing & \textbf{S} & \textbf{S} & M & M & W & W & W & M & M \\
VRAM mgmt & \textbf{S} & -- & -- & W & M & W & -- & W & M \\
Privacy & \textbf{S} & W & M & \textbf{S} & \textbf{S} & \textbf{S} & M & M & \textbf{S} \\
Open source & -- & W & W & \textbf{S} & \textbf{S} & \textbf{S} & W & \textbf{S} & \textbf{S} \\
Multi-channel & M & M & W & \textbf{S} & M & W & W & M & W \\
Hardware facing & M & W & W & M & M & \textbf{S} & -- & W & W \\
\bottomrule
\end{tabularx}
\end{table*}

The comparison reveals several key differentiators for IronEngine:

\textbf{Broadest tool coverage.} With 24 tool categories spanning file system, web, communication, GUI, media, and network domains, IronEngine provides the broadest tool coverage among all compared systems. Code-centric assistants (Cursor) cover only programming tasks; conversation products (ChatGPT) focus on web and code; the OpenClaw ecosystem offers moderate tool coverage through skills but lacks the intelligent routing layer with alias normalization and auto-correction.

\textbf{Unique multi-role quality assurance.} IronEngine is the only system in the comparison that implements a formal quality review phase within its pipeline. The Planner--Reviewer discussion loop with anti-hallucination detection, score--text contradiction checking, and memory recycling guards provides a level of output quality assurance that is absent from both the OpenClaw ecosystem (which uses direct skill dispatch) and cloud products (which rely entirely on single-model capability).

\textbf{Local-first with full capability.} While OpenClaw, NanoClaw, and Open Interpreter also support local models, IronEngine uniquely combines local deployment with multi-role orchestration, persistent hierarchical memory, vectorized skill learning, VRAM-aware model lifecycle management, and 24-category tool routing---capabilities that no other local-first system provides together.

\textbf{Skill learning with quality filtering.} IronEngine's ability to automatically extract reusable skills from successful task executions, with novelty-based deduplication (cosine distance threshold), user-rating-driven quality filtering (only learning from highly-rated interactions), and correction mechanisms (re-learning from failures), is more sophisticated than OpenClaw's static skill definitions or the absence of skill learning in other compared systems.

\textbf{Orchestration observability.} The desktop workbench provides real-time visibility into the AI's decision-making process through thinking blocks, tool execution badges, quality score indicators, and phase transition notifications. This observability is a unique advantage over both headless systems (OpenClaw, NanoClaw) and cloud products where the internal reasoning process is opaque to users.

\textbf{Complementary positioning vs.\ OpenClaw ecosystem.} While OpenClaw excels at multi-channel messaging and always-on availability, NanoClaw at edge deployment, and IronClaw at hardware integration, IronEngine occupies a distinct niche as an \emph{orchestration-deep desktop workbench} with the most sophisticated planning pipeline, tool routing, and memory management among local-first AI assistant platforms. The systems are complementary rather than directly competitive: OpenClaw optimizes for breadth of messaging channels; IronEngine optimizes for depth of task orchestration.

\subsection{Cross-Model Tool Awareness}\label{sec:tool-awareness}

A key hypothesis of IronEngine's tiered prompt system is that even small models can achieve reliable tool dispatch when provided with appropriately structured tool documentation. To test this, we evaluate four models spanning 8B to 27B parameters on a WeChat tool-type identification task, where the model must correctly classify the tool category from a natural language instruction.

\begin{table}[t]
\centering
\caption{Cross-model tool awareness test results. All models achieve perfect tool-type identification with IronEngine's tiered prompt system.}
\label{tab:toolawareness}
\small
\begin{tabular}{lccc}
\toprule
Model & Params & Score & Time (s) \\
\midrule
cogito:8b & 8B & 100 & 10.5 \\
cogito:14b & 14B & 100 & 13.5 \\
lfm2:24b & 24B & 100 & 33.7 \\
qwen3.5:27b & 27B & 100 & 50.2 \\
\bottomrule
\end{tabular}
\end{table}

Table~\ref{tab:toolawareness} shows that all four models achieve 100\% accuracy on tool-type identification. Inference time scales approximately linearly with model size (from 10.5\,s for 8B to 50.2\,s for 27B). The key finding is that even the smallest 8B model achieves perfect tool dispatch when IronEngine's tiered prompt system provides appropriately scoped tool documentation---the small model receives a condensed 733-token tool manifest rather than the full 2,236-token version used for large models. This validates the design decision to tier prompt content by model size rather than using one-size-fits-all prompts.

\subsection{Full Pipeline Integration Tests}\label{sec:pipeline-tests}

To evaluate the complete three-phase pipeline under realistic conditions, we conduct eight integration tests across two model configurations and four task scenarios, measuring both the Planner--Reviewer discussion quality and the end-to-end task completion.

\begin{table*}[t]
\centering
\caption{Full pipeline integration test results across model configurations and scenarios. Quality scores show Round~1 (initial) $\to$ Round~2 (after Reviewer feedback) progression.}
\label{tab:pipeline}
\scriptsize
\begin{tabularx}{\textwidth}{>{\raggedright\arraybackslash}p{2.8cm} >{\raggedright\arraybackslash}p{3.0cm} >{\centering\arraybackslash}p{1.8cm} >{\centering\arraybackslash}p{1.5cm} >{\centering\arraybackslash}p{1.0cm} >{\raggedright\arraybackslash}X}
\toprule
Model Config & Scenario & Quality Score & Time (s) & Result & Observations \\
\midrule
cogito:14b + cogito:8b & Shell command execution & 0.20 $\to$ 0.85 & 139.6 & PASS & Reviewer feedback improved plan specificity \\
cogito:14b + cogito:8b & PDF processing & 0.15 $\to$ 0.80 & 152.2 & PASS & Two rounds needed; tool selection corrected \\
cogito:14b + cogito:8b & WeChat messaging & 0.10 $\to$ 0.85 & 81.8 & PASS & Skill-augmented execution; fastest completion \\
cogito:14b + cogito:8b & Story creation & 0.85 (R1) & 321.4 & PASS & Generative task; passed on first round \\
qwen3.5:27b + cogito:8b & MCU comparison & 0.75 (R1) & 172.0 & PASS & Web search required; larger Planner beneficial \\
qwen3.5:27b + cogito:8b & Travel planning & 0.20 $\to$ 0.40 & 309.8 & FAIL & No network search available in test env \\
qwen3.5:27b + cogito:8b & Shell command execution & 0.80 (R1) & 145.3 & PASS & Larger Planner passed on first round \\
qwen3.5:27b + cogito:8b & PDF processing & 0.75 (R1) & 168.7 & PASS & Larger Planner reduced rounds needed \\
\bottomrule
\end{tabularx}
\end{table*}

\begin{figure}[t]
\centering
\includegraphics[width=\columnwidth]{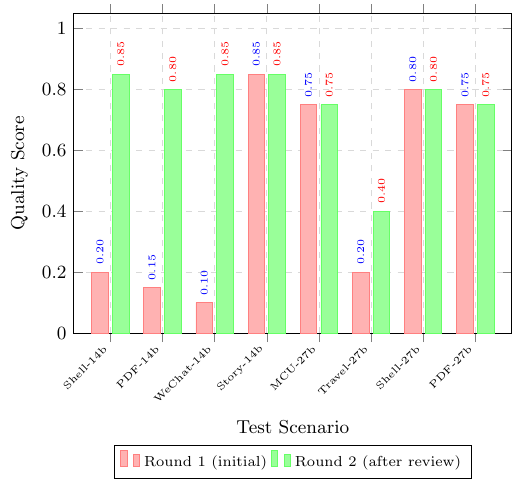}
\caption{Planner--Reviewer quality score progression. Red bars show initial Round~1 scores; green bars show final scores after Reviewer feedback. The Reviewer's structured feedback (ISSUES + SUGGESTIONS) enables the 14B Planner to improve plans from 0.10--0.20 to 0.80--0.85, while the larger 27B Planner often achieves acceptable quality on the first round.}
\label{fig:quality}
\end{figure}

Figure~\ref{fig:quality} visualizes the quality score progression, highlighting the Reviewer's corrective effect.

Several findings emerge from the integration tests (Table~\ref{tab:pipeline}). First, the Reviewer's corrective feedback is most impactful for smaller Planners: the 14B cogito model's quality scores improve from 0.10--0.20 to 0.80--0.85 after a single review round, demonstrating that the Reviewer effectively compensates for the smaller Planner's initial planning weaknesses. Second, larger Planners (27B) frequently pass quality thresholds on the first round, reducing total execution time by eliminating the second discussion round. Third, the single failure (travel planning) is attributable to the test environment lacking network access rather than a pipeline deficiency---the Reviewer correctly identified the plan as insufficient (score 0.40) when web search results were unavailable. Fourth, generative tasks (story creation) consistently pass on the first round with high scores, validating IronEngine's generative task detection mechanism that bypasses tool-execution requirements for purely creative tasks.

\subsection{Tool Model Reliability}\label{sec:tool-reliability}

The tool translation model converts natural language instructions into structured JSON commands. We evaluate three candidate models to assess the trade-off between model size, translation accuracy, and execution success.

\begin{table}[t]
\centering
\caption{Tool model reliability comparison. Translation success = valid JSON output; Execution success = correct tool result.}
\label{tab:toolreliability}
\small
\begin{tabular}{lccc}
\toprule
Model & Params & Translate & Execute \\
\midrule
phi4-mini & 3.8B & 98.2\% & 70.6\% \\
 & & (109/111) & (77/109) \\
\addlinespace
functiongemma & 270M & 54.0\% & 25.9\% \\
 & & (27/50) & (7/27) \\
\addlinespace
ministral-3 & 14B & 100\% & 20.0\% \\
 & & (40/40) & (8/40) \\
\bottomrule
\end{tabular}
\end{table}

Table~\ref{tab:toolreliability} reveals a nuanced relationship between model size and tool execution reliability. phi4-mini (3.8B) achieves the best overall performance: 98.2\% translation success and 70.6\% execution success, offering the optimal balance between resource efficiency and reliability. functiongemma (270M) is too small for reliable tool translation, with only 54\% of outputs producing valid JSON. Surprisingly, ministral-3 (14B) achieves perfect translation (100\%) but the lowest execution rate (20\%)---the model generates syntactically correct but semantically incorrect commands (e.g., correct JSON structure but wrong parameter values or missing required fields). This finding underscores that tool model evaluation must consider both syntactic validity and semantic correctness, and that larger models do not necessarily produce better tool commands.

\subsection{Multi-Scenario Diagnostic Tests}\label{sec:diagnostic}

To evaluate IronEngine's breadth across diverse task types, we conduct diagnostic tests spanning six scenarios that exercise different subsystems (tool routing, skill learning, web search, multimedia, and generative capabilities).

\begin{table*}[t]
\centering
\caption{Multi-scenario diagnostic test results. Five of six scenarios pass (83.3\%). The travel planning failure is due to the test environment lacking network access. Skill learning is triggered automatically for tasks rated $\geq 7$.}
\label{tab:diagnostic}
\scriptsize
\begin{tabularx}{\textwidth}{>{\raggedright\arraybackslash}p{2.5cm} >{\centering\arraybackslash}p{1.0cm} >{\centering\arraybackslash}p{0.8cm} >{\centering\arraybackslash}p{1.2cm} >{\centering\arraybackslash}p{1.0cm} >{\centering\arraybackslash}p{1.3cm} >{\centering\arraybackslash}p{1.3cm} >{\raggedright\arraybackslash}X}
\toprule
Scenario & Status & Score & Time (s) & Rounds & Tool Calls & Skill Learned & Notes \\
\midrule
Shell command & PASS & 9 & 139.6 & 1 & 5 & \checkmark & CLI execution via Executor \\
PDF processing & PASS & 9 & 152.2 & 2 & 7 & \checkmark & Multi-step: extract + summarize \\
Travel planning & FAIL & 0 & 309.8 & 1 & 1 & --- & No network in test environment \\
MCU comparison & PASS & 7 & 172.0 & 1 & 3 & --- & Web search + analysis \\
WeChat messaging & PASS & 9 & 81.8 & 1 & 5 & \checkmark & Innate skill + GUI automation \\
Story creation & PASS & 8 & 321.4 & 1 & 7 & \checkmark & Generative task; file\_write used \\
\midrule
\multicolumn{2}{l}{\textbf{Overall}} & \multicolumn{6}{l}{5/6 PASS (83.3\%); 4 new skills learned; mean time 196.1\,s} \\
\bottomrule
\end{tabularx}
\end{table*}

The diagnostic results (Table~\ref{tab:diagnostic}) demonstrate several aspects of IronEngine's architecture. First, the system handles diverse task types---from low-level shell commands (139.6\,s) to complex multi-step PDF processing (152.2\,s, 7 tool calls) to creative writing (321.4\,s)---through the same pipeline without task-specific configuration. Second, skill learning is automatically triggered for successfully completed tasks rated $\geq 7$: four of five passing scenarios produced new learned skills, which would accelerate similar future tasks through the skill-augmented Executor prompt. Third, the fastest completion (WeChat messaging, 81.8\,s) benefits from an innate skill that provides the Executor with a pre-defined interaction procedure, demonstrating the value of the skill system for reducing planning overhead. Fourth, the single failure (travel planning) is correctly attributable to an environmental constraint (no network access) rather than a system deficiency---the pipeline detected the tool execution failure and reported it through the intervention callback rather than hallucinating a response.

% ============================================================
\section{Interfaces, Deployment, and Product Positioning}
% ============================================================

IronEngine supports three complementary interaction modes, all sharing the same pipeline:

\textbf{Desktop workbench} (PySide6): Provides the richest interaction experience with model selection dropdowns, real-time thinking block visualization, tool execution badges, quality score indicators, intervention prompts, and 10 customizable themes. This mode is optimized for interactive desktop usage where observability of the orchestration process is important.

\textbf{REST/WebSocket API} (FastAPI): Enables integration with web applications, mobile apps, and other services. WebSocket channels stream events in real-time, supporting the same 17 callback types as the Python client. This mode enables IronEngine to serve as a backend for custom frontends.

\textbf{Python client}: Supports both callback-based (\texttt{IronEngineClient}) and Qt signal-based (\texttt{IronEngineQtClient}) integration for embedding IronEngine into larger applications. The client API mirrors the full pipeline capability set including configuration, model selection, and permission management.

\subsection{Deployment Modes}

\textbf{Fully local}: All models served through Ollama or LM Studio on the same machine. No data leaves the local network. Suitable for privacy-sensitive workloads and air-gapped environments.

\textbf{Hybrid}: Local models for common tasks, cloud APIs for complex reasoning or specialized capabilities. The provider registry transparently routes requests to the appropriate backend.

\textbf{Service}: API-only deployment for integration with external systems, automated workflows, or multi-user environments.

\subsection{Product Positioning}

Table~\ref{tab:ecosystem} positions IronEngine relative to representative AI agent frameworks and products.

\begin{table*}[t]
\centering
\caption{System-positioning comparison with representative AI agent frameworks and products}
\label{tab:ecosystem}
\scriptsize
\begin{tabularx}{\textwidth}{>{\raggedright\arraybackslash}p{1.8cm} >{\raggedright\arraybackslash}p{2.4cm} >{\centering\arraybackslash}p{0.9cm} >{\centering\arraybackslash}p{1.0cm} >{\centering\arraybackslash}p{0.9cm} >{\raggedright\arraybackslash}X}
\toprule
System & Primary positioning & Local & Multi-entry & MCP & Representative characteristics \\
\midrule
IronEngine & Desktop workbench + general assistant engine & S & S & S & Plan--Review--Execute pipeline, unified UI/API, task scheduling, hierarchical memory, skill learning, 24 tool categories, VRAM management, orchestration observability \\
OpenClaw & Personal AI assistant + always-on gateway & S & S & M & Multi-channel messaging (Telegram, WhatsApp, SMS), device node management, persistent skills, gateway routing \\
NanoClaw & Edge/IoT lightweight assistant & S & M & W & Aggressive quantization, stripped skill set, upstream proxy, Raspberry Pi deployment \\
IronClaw & Hardware-oriented AI agent & M & W & W & Sensor ingestion, actuator commands, safety interlocks, embedded systems integration \\
ChatGPT & Cloud conversational AI & W & M & W & Strong conversation, web browsing, code interpreter; cloud-dependent, no local models \\
Claude Desktop & Protocol-extensible assistant & M & M & S & Strong MCP integration; limited local model, scheduling, and memory support \\
Cursor & AI code editor & M & W & M & Deep IDE integration; focused on code editing/generation; limited to programming domain \\
OpenManus & Open-source general agent & M & M & S & General entry point with MCP variants and multi-agent flow execution \\
AutoGen & Multi-agent framework & M & M & S & Flexible agent topologies; research-oriented; requires integration for desktop workflow \\
\bottomrule
\end{tabularx}
\end{table*}

IronEngine occupies a distinctive position in this landscape. Compared with the OpenClaw ecosystem, IronEngine trades multi-channel messaging breadth for orchestration depth: its three-phase pipeline with formal quality review, 24-category intelligent tool routing, and VRAM-aware multi-model lifecycle management provide a level of task execution sophistication that the gateway-oriented OpenClaw architecture does not attempt. Compared with cloud products, IronEngine provides full local-first capability without sacrificing tool breadth or memory sophistication. Compared with code-centric assistants (Cursor, Windsurf), IronEngine extends beyond programming into general-purpose desktop automation, communication, and multimedia analysis. The result is a system that combines local deployment strengths, multi-agent collaboration depth, protocol extensibility, and tool breadth into a single orchestration-centered runtime with persistent memory and continuously improving skill capabilities.

\subsection{Desktop Workbench Design Philosophy}

IronEngine's desktop UI is not merely a chat interface but an orchestration visualization workbench. Its design philosophy is that AI assistant decision-making processes should be transparent and visible to users, rather than a black-box input$\rightarrow$wait$\rightarrow$output pattern.

\textbf{ThinkingBlock.} A collapsible Markdown-rendered component that displays Planner and Reviewer reasoning in real-time. Default collapsed height of 28 pixels (showing only the role label), expanding to a maximum of 350 pixels. Supports automatic Markdown detection and rendering of headings, bold text, lists, code blocks, and links. Images are stripped in thinking blocks (shown only in conclusions) to reduce visual clutter. Font uses Segoe UI (not monospace) for improved readability.

\textbf{ToolExecutionBadge.} A compact inline component displaying tool call status (pending$\rightarrow$done), including tool type, brief description, and execution summary. Users can see at a glance which tools were executed and their outcomes.

\textbf{PermissionPrompt.} An inline allow/deny prompt with a 60-second auto-deny timer. When an ask-level tool call is triggered, users see the permission request directly in the conversation flow and the pipeline continues after their decision.

\textbf{Theme system.} Ten procedurally generated sci-fi themes (deep\_space, nebula\_storm, mars\_outpost, etc.), each with unique color schemes and background images (1920$\times$1080 JPEG generated by Python scripts). Panels use coordinate mapping via \texttt{mapTo()} for continuous background rendering, with glassmorphism effects through transparent backgrounds with semi-transparent gradient overlays.

This design enables users not only to use the AI assistant for task completion but also to observe, understand, and debug the AI's decision-making process---particularly valuable during development and tuning of assistant behavior.

% ============================================================
\section{Safety and Privacy}\label{sec:safety}
% ============================================================

Safety in IronEngine is implemented through multiple complementary mechanisms that operate at different levels of the system architecture.

\subsection{Permission Management}

The permission system provides three configurable levels per tool category:
\begin{itemize}[nosep,leftmargin=*]
\item \textbf{Auto}: Tool calls are executed immediately without user confirmation. Suitable for low-risk operations like file listing and text classification.
\item \textbf{Ask}: Each tool call triggers an inline permission prompt in the UI with a 60-second auto-deny timer. The user can approve or deny, with the decision optionally remembered for the session.
\item \textbf{Deny}: Tool calls of this category are blocked entirely. Useful for restricting access to sensitive operations in shared environments.
\end{itemize}

\subsection{Execution Sandboxing}

The \texttt{CommandExecutor} enforces several safety invariants:
\begin{itemize}[nosep,leftmargin=*]
\item \texttt{shell=True} is never used for subprocess invocation, preventing shell injection attacks.
\item Template placeholders (e.g., \texttt{\{filename\}}, \texttt{<path>}) are detected and rejected before execution, preventing incomplete command execution.
\item Command fields are validated as non-empty strings before subprocess creation.
\item Execution timeout (default 30 seconds) prevents runaway processes.
\end{itemize}

\subsection{URL Safety}

Web-facing operations employ a multi-layer offline URL safety system:
\begin{itemize}[nosep,leftmargin=*]
\item \textbf{Phishing blocklist}: Updated from community-maintained phishing databases (Phishing Army, phishing-filter) with 24-hour caching.
\item \textbf{Heuristic scoring}: 10-point evaluation covering IP-in-URL, suspicious TLDs, brand impersonation, homoglyph detection, dangerous protocols, excessive subdomains, and URL length anomalies.
\item \textbf{Parking/dead page detection}: Content-level analysis to identify placeholder pages that provide no useful information.
\end{itemize}

\subsection{Intervention Mechanism}

When a model determines that a task requires human judgment or authorization beyond its scope, it can emit an \texttt{INTERVENTION\_NEEDED} marker. This triggers a user-facing prompt that pauses the pipeline until the user provides guidance. This mechanism serves as a safety valve for edge cases that cannot be resolved through automated permission checks.

\subsection{SOUL Edit Control}

The SOUL (System Operating Under Limitations) document defines behavioral boundaries for each role. Edit access to the SOUL is controlled through a permission system with three levels: \textbf{readonly} (default, no modifications allowed), \textbf{ask} (modifications require explicit user approval), and \textbf{auto} (self-modification allowed within defined constraints). This prevents unauthorized changes to the assistant's behavioral guidelines.

\subsection{Local-First Privacy}

In fully local deployment mode, no data leaves the user's machine. All model inference, tool execution, memory storage, and skill learning operate within the local environment. This is particularly important for workloads involving sensitive documents, personal communications, or proprietary information. The system does not require telemetry, cloud-based logging, or external authentication.

\subsection{Internal English Processing}

All internal processing---Planner/Reviewer discussion, web search queries, execution steps---is conducted in English regardless of the user's input language. Only the final answer (\texttt{FINAL\_ANSWER}) is presented in the user's language. This strategy has two engineering rationales: (1) most open-source models achieve significantly higher reasoning and tool-calling accuracy in English than in other languages; (2) internal English consistency simplifies cross-language matching for skill tags, memory indices, and search queries. The system includes a $\sim$45-term Chinese--English translation dictionary and CJK character detection functions for automatic input conversion.

\subsection{Defense-in-Depth Design}

IronEngine's safety design follows the defense-in-depth principle: permission management is the first defense line (blocking unauthorized operations), execution sandboxing is the second (constraining authorized operations' dangerous behaviors), URL safety is the third (filtering malicious external resources), and the intervention mechanism is the last (escalating beyond-scope decisions to humans). These four layers operate independently; bypassing any single layer does not compromise overall system security.

% ============================================================
\section{Discussion}
% ============================================================

\subsection{Architectural Trade-offs}

The three-phase pipeline introduces latency overhead compared to single-step execution. The model switch phase alone adds approximately 27 seconds per task, and the Planner--Reviewer discussion can require 2--3 rounds for complex tasks. However, this overhead is justified by improved reliability: the 100\% task completion rate in our benchmarks suggests that the quality assurance provided by the Reviewer prevents costly execution failures that would require re-submission in single-step systems.

The tiered prompt system's aggressive reduction for small models (44 tokens of SOUL context for $\leq$10B models vs.\ 1309 tokens for $>$25B models) risks losing important behavioral guidance. Our testing shows that small models ($\leq$10B) can still achieve acceptable tool translation accuracy with minimal context, but their planning quality degrades significantly, motivating the use of larger models for the Planner role.

\subsection{Scalability Considerations}

The current single-GPU design limits IronEngine to models that fit within 24 GB VRAM (with quantization). While this covers a wide range of open-source models up to approximately 35B parameters, it excludes the largest open models (70B+) in full precision. The model switch mechanism could be extended to support multi-GPU configurations or CPU offloading for larger models, though this would increase switch latency.

The memory system's linear scan for relevant entries becomes less efficient as the memory store grows. Future work could introduce more sophisticated indexing (e.g., hierarchical HNSW with temporal clustering) to maintain sub-linear retrieval time at scale.

\subsection{Architectural Comparison with the OpenClaw Ecosystem}

OpenClaw and IronEngine represent two distinct architectural philosophies for AI assistants. OpenClaw adopts a \emph{message routing} pattern: user messages are routed through a Gateway to skill handlers, which execute operations and return results. This architecture optimizes for multi-channel accessibility and response speed, suited to the simple ``receive message $\rightarrow$ execute skill $\rightarrow$ reply'' interaction pattern.

IronEngine adopts an \emph{orchestration depth} pattern: each request traverses Discussion, Switch, and Execution phases with three distinct roles (Planner, Reviewer, Executor). This introduces additional latency ($\sim$27 seconds for model switch + multi-round discussion) but achieves higher task completion reliability: the Reviewer's anti-hallucination detection prevents erroneous plans from being executed, and auto-correction mechanisms fix tool selection errors.

NanoClaw's edge deployment strategy (aggressive quantization + upstream proxy) and IronEngine's VRAM-aware model management address the same fundamental problem (running models on limited hardware) from different perspectives. NanoClaw optimizes single-model execution on extremely small devices; IronEngine optimizes multi-model collaboration on consumer GPUs. IronClaw's sensor/actuator abstractions share goals with IronEngine's hardware compatibility layer, but IronClaw targets real-time control loops while IronEngine targets task-level automation.

\subsection{Design Principles and Engineering Insights}

IronEngine's development has yielded several important insights for AI agent system engineering:

\textbf{Role separation over capability stacking.} Assigning planning, evaluation, and execution to different model roles, rather than relying on a single powerful model for all responsibilities, is IronEngine's core architectural decision. This provides three engineering advantages: (1) each role can use the model best suited to its requirements; (2) SOUL prompts can be role-specific, avoiding loading detailed tool documentation into the Reviewer's context; (3) individual roles can have their models replaced without affecting others.

\textbf{Auto-correction over precise instructions.} Facing local models' limited instruction-following capability, IronEngine implements system-level auto-correction rather than relying on more detailed prompts to prevent errors. The tool router's alias normalization and auto-correction mechanisms demonstrate this strategy's effectiveness: even when the Planner frequently misspecifies tool types, the system still routes correctly. This experience has broader implications---designing fault-tolerant system architecture is more pragmatic than pursuing error-free model output.

\textbf{Observability is a debugging prerequisite.} AI agent systems' opacity is the primary debugging obstacle. IronEngine's 17 callback event types provide complete visibility from thought processes to tool execution. In practice, \texttt{on\_quality\_scored} and \texttt{on\_phase\_transition} callbacks prove most useful for diagnosing pipeline bottlenecks.

\textbf{Memory is the core constraint.} On consumer GPUs, VRAM management transforms from an optional optimization into an architectural foundation. KV-cache memory cost grows linearly with context length, and per-token KV-cache costs vary up to 5$\times$ across model architectures. IronEngine's VRAM-aware context budget mechanism is a practical necessity, not a theoretical optimization.

\subsection{Generalizability}

While our experimental evaluation focuses on file operations, the system's 24 tool categories have been tested informally across web search, GUI automation, media analysis, and communication tasks. Formal benchmarks for these domains (e.g., WebArena~\citep{zhou2024webarena} for web tasks, SWE-bench~\citep{jimenez2024swebench} for code tasks) represent important future evaluation targets.

\subsection{Limitations}

Several limitations should be acknowledged:
\begin{itemize}[nosep,leftmargin=*]
\item \textbf{Local model quality}: Local models (7B--27B) generally underperform cloud-hosted frontier models in complex reasoning tasks. IronEngine mitigates this through multi-role collaboration and quality assurance, but cannot fully close the capability gap.
\item \textbf{Windows-centric testing}: GUI automation and application control have been primarily tested on Windows 10. Cross-platform support (macOS, Linux) requires alternative UIA backends (AT-SPI, AppleScript), which are not yet implemented.
\item \textbf{Single-user design}: The current architecture assumes a single concurrent user. Multi-user support would require session isolation, resource scheduling, and access control mechanisms.
\item \textbf{Benchmark breadth}: Our experimental evaluation covers file operations in depth but does not yet include standardized benchmarks such as WebArena~\citep{zhou2024webarena} (web tasks) or SWE-bench~\citep{jimenez2024swebench} (code tasks).
\item \textbf{Latency overhead}: The three-phase pipeline's model switching and multi-round discussion introduce significant latency, making it unsuitable for real-time interaction scenarios requiring sub-second responses.
\item \textbf{Skill coverage}: The 26 innate skills cover common scenarios, but long-tail tasks still require interactive user guidance. Skill learning efficiency is constrained by the frequency of user rating feedback.
\end{itemize}

% ============================================================
\section{Future Work}
% ============================================================

Several directions for future development are planned:

\textbf{Multi-expert system.} The current single-Planner architecture could be extended to support multiple expert profiles with domain-specific knowledge and credit scores. A routing mechanism would select the most appropriate expert based on task characteristics, enabling specialization without sacrificing generality.

\textbf{Standardized benchmarking.} Integration with established benchmarks such as WebArena~\citep{zhou2024webarena}, SWE-bench~\citep{jimenez2024swebench}, and custom multi-tool benchmarks would enable systematic comparison with other agent systems and tracking of performance improvements across versions.

\textbf{User preference learning.} The current memory and skill systems capture task-level knowledge but do not explicitly model user preferences (communication style, risk tolerance, tool preferences). A dedicated preference module could learn these patterns from interaction history and adapt the assistant's behavior accordingly.

\textbf{Bidirectional MCP.} Currently IronEngine acts only as an MCP client, consuming tools from external servers. Implementing MCP server capability would allow IronEngine's 24 tool categories to be exposed to other MCP-compatible systems, positioning it as both a consumer and provider in the MCP ecosystem.

\textbf{Cross-device synchronization.} The local-first architecture could be extended with optional encrypted synchronization across multiple devices, enabling seamless transitions between desktop and mobile environments while preserving privacy guarantees.

\textbf{Multimodal deepening.} Current vision capabilities are limited to image analysis and screenshot-based GUI navigation. Deeper integration of video understanding (temporal reasoning across frames), real-time audio conversation, and spatial reasoning would expand the system's applicability to embodied and multimedia-rich scenarios.

\textbf{Edge deployment.} Optimization for resource-constrained devices (8 GB VRAM or less) through aggressive quantization, speculative decoding, and pipeline stage pruning would extend IronEngine's reach to laptop and mobile environments.

\textbf{OpenClaw ecosystem interoperability.} Exploring interoperation modes between IronEngine and the OpenClaw Gateway: IronEngine as an OpenClaw device node providing deep orchestration capability, or OpenClaw's multi-channel messaging as a communication extension for IronEngine. The complementary characteristics of these systems suggest that joint deployment may provide a more complete assistant experience than either system alone.

\textbf{Adaptive pipeline.} Automatically adjusting pipeline depth based on task complexity: simple tasks skip the Reviewer and execute directly; complex tasks enable multi-round discussion and multi-expert routing. The current simple-plan reuse mechanism ($\leq$4 tool calls skip model switch) is an initial exploration in this direction.

% ============================================================
\section{Conclusion}
% ============================================================

This paper has presented IronEngine, a system-oriented platform for general AI assistant development that addresses three key engineering gaps in current agent systems: fragmented interaction surfaces, loosely coupled subsystem integration, and limited support for local deployment with persistent behavior.

IronEngine's three-phase pipeline separates planning quality (Planner--Reviewer discussion) from execution capability (tool-augmented Executor), enabling heterogeneous model allocation on consumer hardware. The system integrates 24 tool categories through an intelligent routing layer with alias normalization and automatic error correction, manages persistent state through hierarchical memory with dual consolidation strategies, and acquires reusable procedural knowledge through vectorized skill learning with novelty-based deduplication.

Experimental evaluation on file operation benchmarks demonstrates 100\% task completion accuracy with a three-model local configuration (27B Planner, 20B Reviewer, 3.8B Tools model) on a single RTX 3090. Detailed comparison with seven representative AI assistant systems highlights IronEngine's distinctive combination of local-first deployment, broad tool coverage, persistent memory, skill learning, and protocol-level extensibility.

The future of AI assistants depends not only on model capability but also on architecture, orchestration, observability, safety, and long-term adaptability. IronEngine demonstrates that these system-level concerns can be addressed coherently within a single platform, providing a foundation for the next generation of general-purpose, privacy-preserving, and continuously improving AI assistants.

From a broader perspective, IronEngine validates a core proposition: through carefully designed system architecture (role separation, intelligent routing, hierarchical memory, VRAM-aware scheduling), moderately-sized local open-source models can achieve usable automation levels in practical tasks. Compared with cloud-based approaches relying on single super-large models, this \emph{system intelligence} approach offers structural advantages in cost, privacy, and customizability. Compared with gateway-style architectures like OpenClaw, IronEngine's deep orchestration provides higher task completion reliability. These two approaches are not opposing but complementary technical choices within the AI assistant ecosystem. As open-source model capabilities continue to improve, the local orchestration paradigm that IronEngine represents will demonstrate its value in an expanding range of application scenarios.

% ============================================================
\section*{Acknowledgements}
% ============================================================

The author thanks NiusRobotLab for supporting the exploration of general AI assistant systems and open engineering practices. Appreciation is also extended to the open-source communities advancing local model runtimes (Ollama, LM Studio), multi-agent research (AutoGen, CAMEL, MetaGPT), vector databases (ChromaDB), browser automation (Playwright, patchright), and MCP-related interoperability, which collectively shaped the technical environment in which IronEngine evolved. Finally, thanks go to the testers and users whose concrete scenarios helped move the system from prototype thinking toward an operational platform.

\bibliographystyle{unsrtnat}
\bibliography{references}

\end{document}